\documentclass[journal,onecolumn,draftcls,12pt,twoside]{IEEEtran}

\usepackage{times}
\usepackage{epsfig}
\usepackage{graphicx}
\usepackage{amsmath}
\usepackage{amssymb}
\usepackage{enumitem} 
\usepackage{graphicx}
\usepackage{subcaption}

\usepackage{cite}

\begin{document}
\title{AMNet: Deep Atrous Multiscale \\ Stereo Disparity Estimation Networks}

\author{Xianzhi Du, Mostafa El-Khamy, Jungwon Lee 
\\ Samsung SOC R\&D Lab, San Diego, CA 92121, USA}

\maketitle

\maketitle

\begin{abstract}
  In this paper, a new deep learning architecture for stereo disparity estimation is proposed. The proposed atrous multiscale network (AMNet) adopts an efficient feature extractor with depthwise-separable convolutions and an extended cost volume that deploys novel stereo matching costs on the deep features. A stacked atrous multiscale network is proposed to aggregate rich multiscale contextual information from the cost volume which allows for estimating the disparity with high accuracy at multiple scales. AMNet can be further modified to be a foreground-background aware network, FBA-AMNet, which is capable of discriminating between the foreground and the background objects in the scene at multiple scales. An iterative multitask learning method is proposed to train FBA-AMNet end-to-end. The proposed disparity estimation networks, AMNet and FBA-AMNet, show accurate disparity estimates and advance the state of the art on the challenging Middlebury, KITTI 2012, KITTI 2015, and Sceneflow stereo disparity estimation benchmarks.

\end{abstract}

\section{Introduction}
\IEEEPARstart{D}{epth} estimation is a fundamental computer vision problem aiming to predict a measure of distance of each point in a captured scene. Accurate depth estimation has many applications such as scene understanding, autonomous driving, computational photography, and improving the aesthetic quality of images by synthesizing the Bokeh effect. Given a rectified stereo image pair, depth estimation can be done by disparity estimation with camera calibration. For each pixel in one image, disparity estimation finds the shifts between one pixel and its corresponding pixel in the other image on the same horizontal line so that the two pixels are the projections of a same 3D position.

\begin{figure}[h!]
    \centering
    \begin{subfigure}[b]{0.3\textwidth}
        \includegraphics[width=\textwidth]{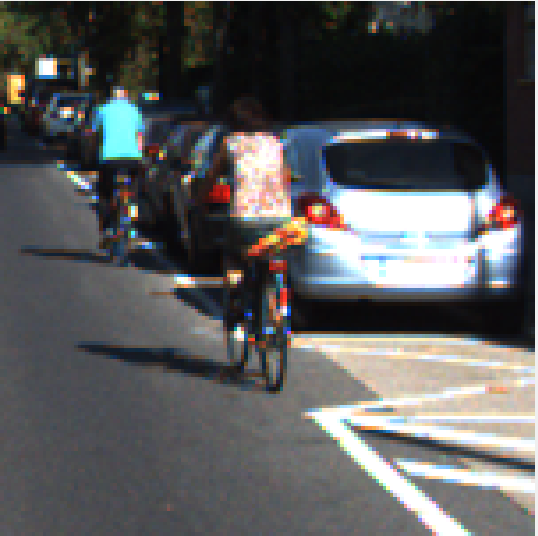}
    \end{subfigure}
    ~ 
    \begin{subfigure}[b]{0.3\textwidth}
        \includegraphics[width=\textwidth]{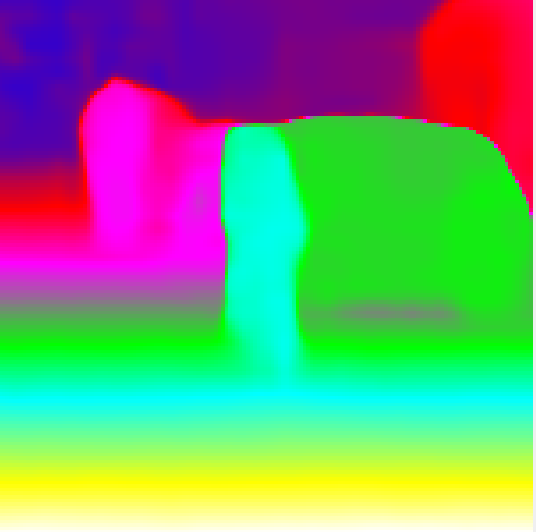}
    \end{subfigure}
    ~ 
    \begin{subfigure}[b]{0.3\textwidth}
        \includegraphics[width=\textwidth]{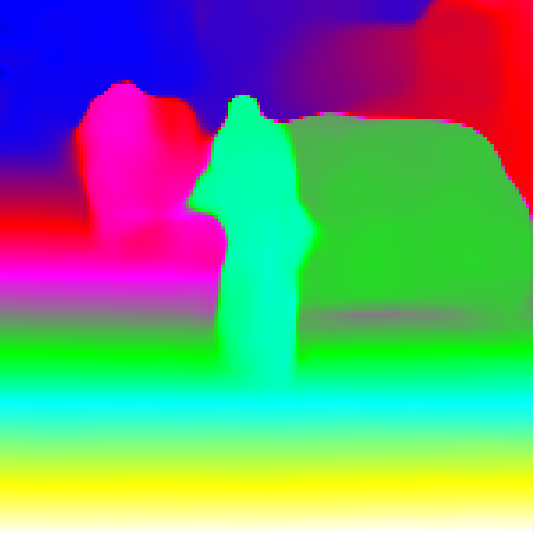}
    \end{subfigure}
    \begin{subfigure}[b]{0.3\textwidth}
        \includegraphics[width=\textwidth]{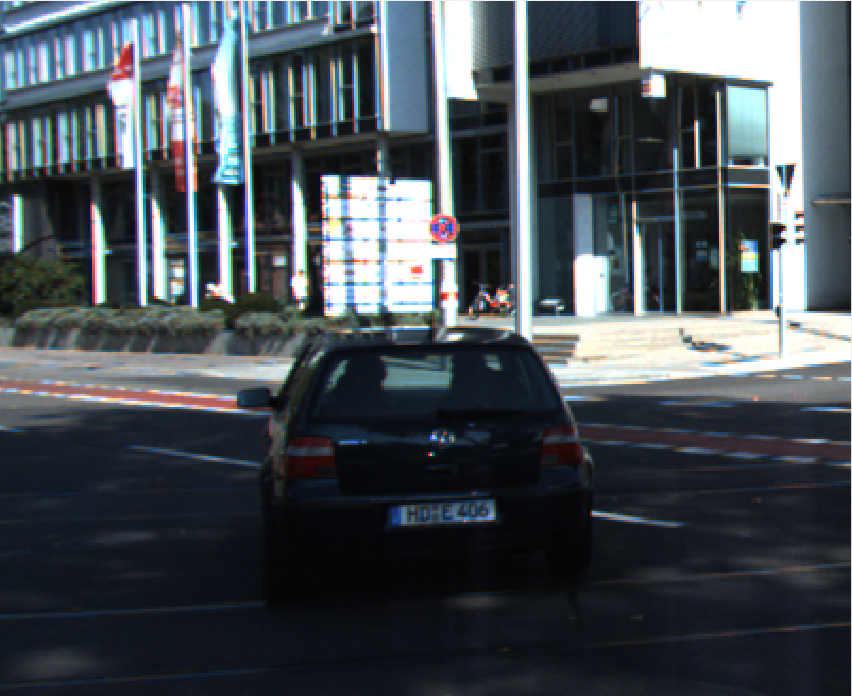}
    \caption{Input image}
    \end{subfigure}
    ~ 
    \begin{subfigure}[b]{0.3\textwidth}
        \includegraphics[width=\textwidth]{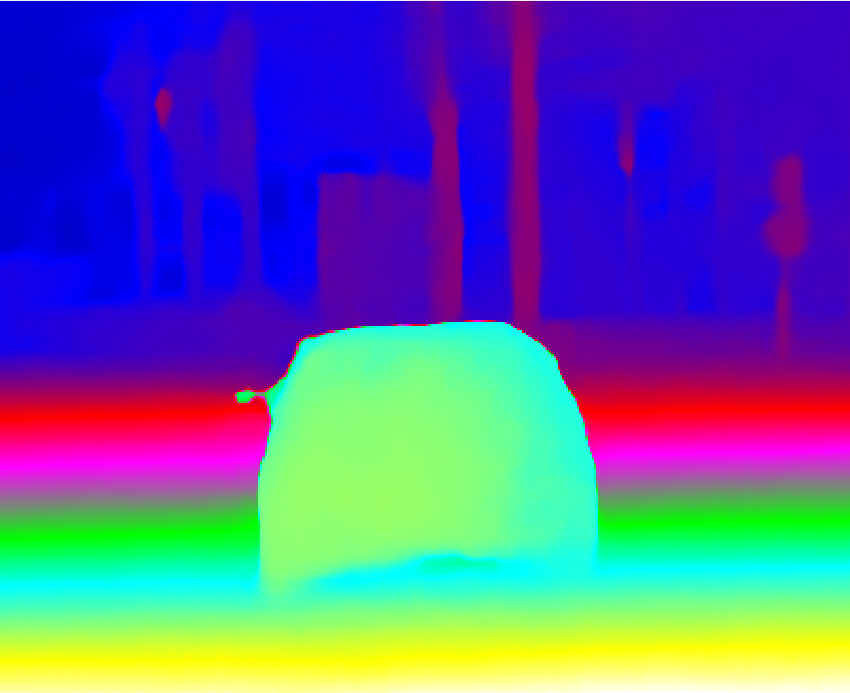}
    \caption{AMNet}
    \end{subfigure}
    ~ 
    \begin{subfigure}[b]{0.3\textwidth}
        \includegraphics[width=\textwidth]{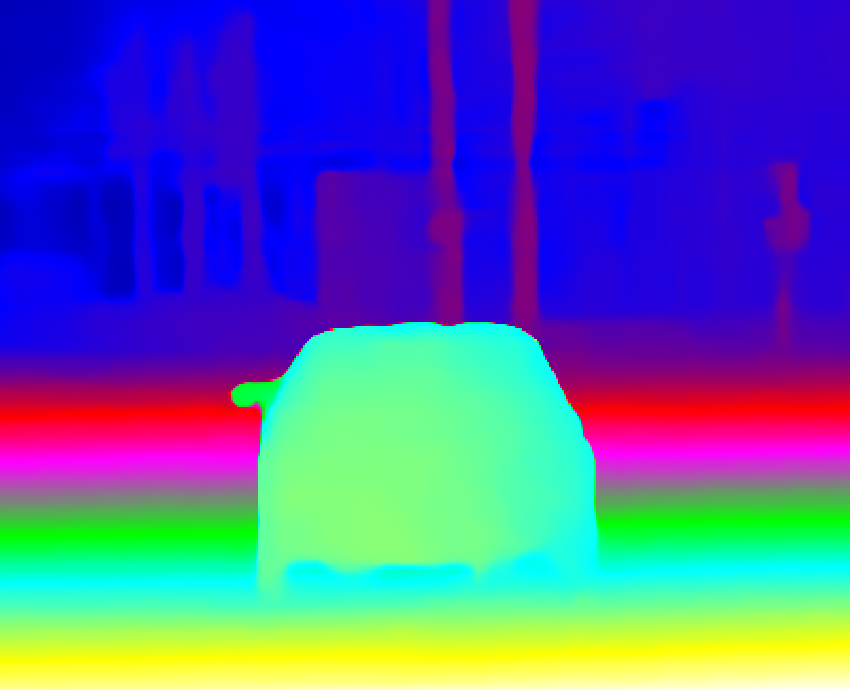}
    \caption{FBA-AMNet}
    \end{subfigure}
\caption{Disparity maps generated by the proposed Atrous Multiscale Network (AMNet-32) and the proposed Foreground-Background Aware AMNet (FBA-AMNet-32) on two challenging foreground objects from KITTI test images.}
\label{fig:fgcomp}
\end{figure}

Disparity estimation based on a stereo image pair is a well known problem in computer vision. Often the stereo images are first rectified to lie in the same image plane and such that corresponding pixels in the left and right lie on the same horizontal line. Disparity estimation pipelines classically consist of three or fours steps; feature extraction, matching cost computation, disparity aggregation and computation, and  an optional disparity refinement step \cite{scharstein2002taxonomy}.  Calculation of the matching cost at a given disparity is based on evaluating a function that measures the similarity between pixels in the left and right images with this disparity shift, which can simply be the sum of absolute differences of pixel intensities at the given disparity \cite{PAMI_SMC}.  Calculation of the matching costs on pixel intensities directly is prone to errors due to practical variations such as illumination differences, inconsistencies, environmental factors such as rain and snow flares,  and occlusions. Hence, the robustness of traditional stereo matching methods can be improved by first extracting features from the intensities such as local binary patterns \cite{guo2010completed} and local dense encoding \cite{LDP}. Disparity aggregation can be done by simple aggregation of the calculated cost over local box windows, or by guided-image cost volume filtering \cite{costvolumefiltering}. Disparity calculation can be done by local, global or semiglobal methods. Semiglobal matching (SGM) \cite{SGM} is considered the most popular method, as it is more robust than local window-based methods and performs cost aggregation by approximate minimization of a two dimensional energy function towards each pixel along eight one dimensional paths. SGM is less complex than global methods such graph cuts (GC) that minimize the two dimensional energy function with a full two dimensional connectivity for the smoothness term \cite{kolmogorov2004energy}. Traditionally, disparity refinement is done by further checking for left and right consistencies, invalidating occlusions and mismatches, and filling such invalid segments by propagating neighboring disparity values.

\begin{figure*}[t!]
\begin{center}
   \includegraphics[width=0.95\linewidth]{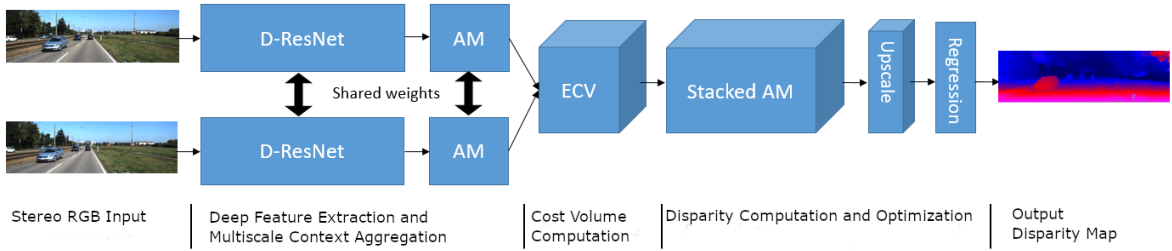}
\end{center}
   \caption{Architecture of proposed Atrous Multiscale Network (AMNet) for stereo disparity estimation.}
\label{fig:amnet}
\end{figure*}

Recently, there has been significant efforts in collecting datasets with stereo input images and their ground truth disparity maps, e.g. SceneFlow~\cite{sceneflow}, KITTI 2012~\cite{kitti2012}, KITTI 2015~\cite{kitti2015}, and the Middlebury~\cite{middlebury} stereo benchmark datasets. The existence of such datasets enabled supervised training of deep neural networks for the task of stereo matching,  as well as the transparent testing and benchmarking of different algorithms on their hosting servers. Convolutional neural networks (CNN) have become ubiquitous in addressing image processing and computer vision  problems. CNN-based disparity estimation systems take their cues from the classical ones, and constitute of different modules that attempt to perform the same four tasks of feature extraction, matching cost estimation, disparity aggregation and computation, and disparity refinement. First, deep features are extracted from the rectified left and right images using deep convolutional networks such as ResNet-50~\cite{resnet} or VGG-16~\cite{vgg}. The cost volume (CV) is formed by measuring the matching cost between the extracted left and  right deep feature maps. Typical choices for the matching cost can be by simple feature concatenation or by calculation of metrics such as absolute distance or correlation~\cite{sceneflow,MCCNN,flownet,luo}. The CV is further processed and refined by a disparity computation module that regresses to the estimated disparity. Refinement networks can then be used to further refine the initial coarse depth or disparity estimates.

\begin{figure}[h!]
    \centering
    \begin{subfigure}[b]{0.4\textwidth}
        \includegraphics[width=\textwidth]{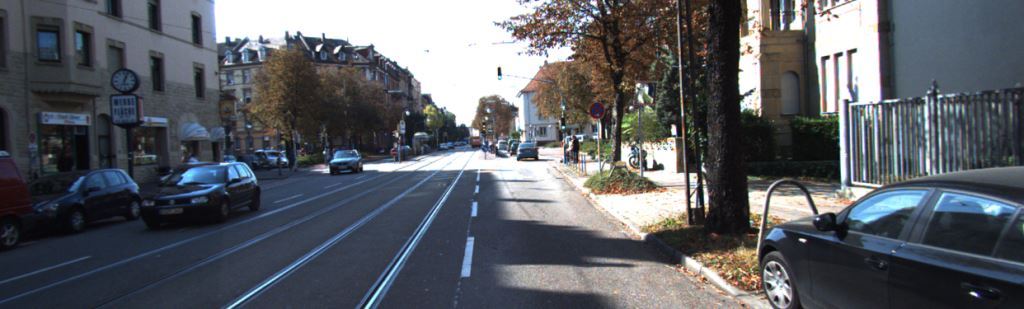}
        \caption{Left input image}
    \end{subfigure}
 \\
    \begin{subfigure}[b]{0.4\textwidth}
        \includegraphics[width=\textwidth]{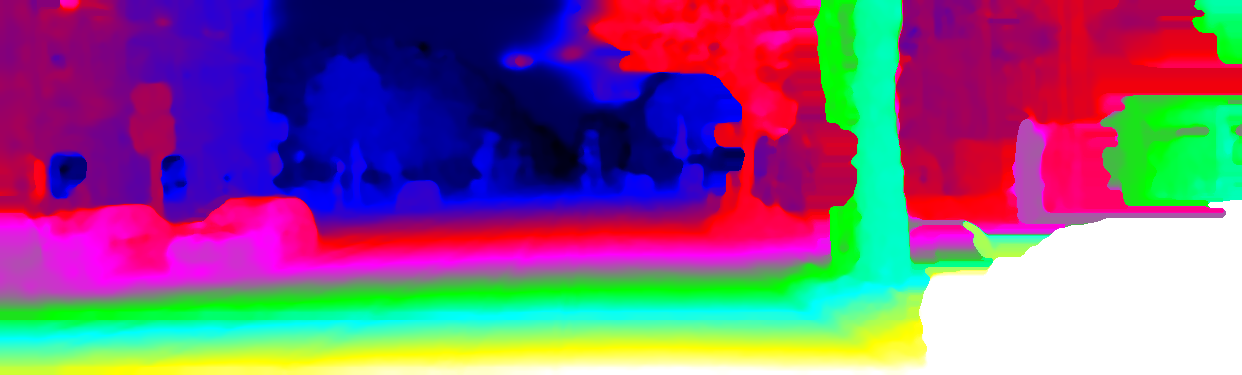}
        \caption{DispSegNet (D1-all = 7.90\%) \cite{DispSegNet} }
    \end{subfigure}
    \qquad
		   \begin{subfigure}[b]{0.4\textwidth}
        \includegraphics[width=\textwidth]{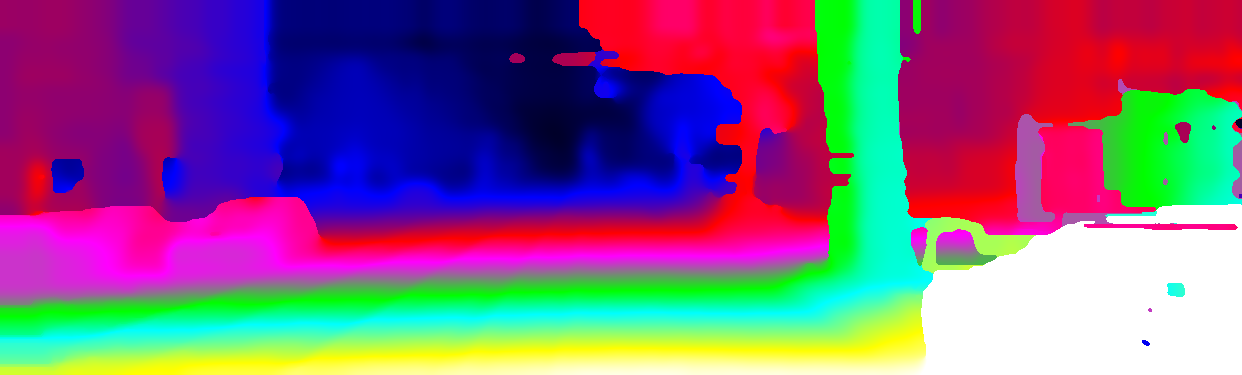}
        \caption{MC-CCN (D1-all = 4.99\%) \cite{MCCNN} }
    \end{subfigure}
	 \\
      \begin{subfigure}[b]{0.4\textwidth}
        \includegraphics[width=\textwidth]{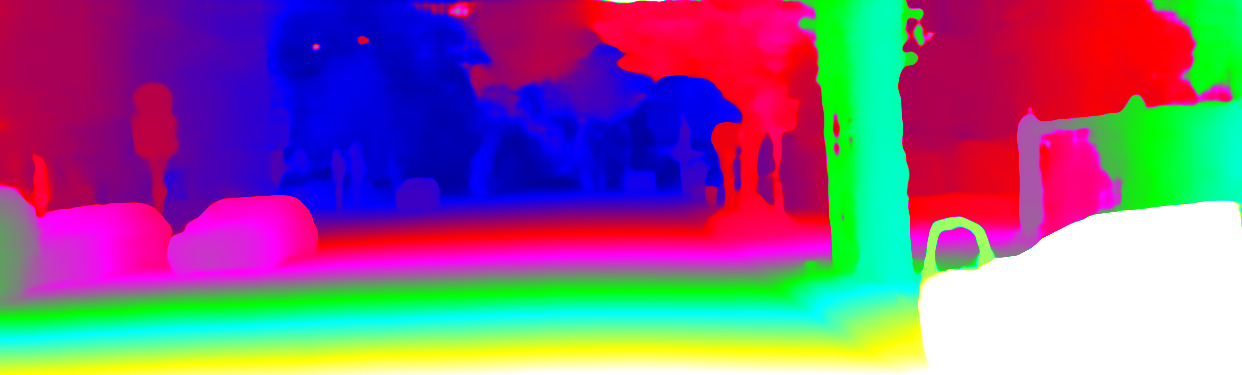}
        \caption{MS-CSPN (D1-all = 2.04\%) \cite{CSPN} }
    \end{subfigure}
     \qquad
		 \begin{subfigure}[b]{0.4\textwidth}
        \includegraphics[width=\textwidth]{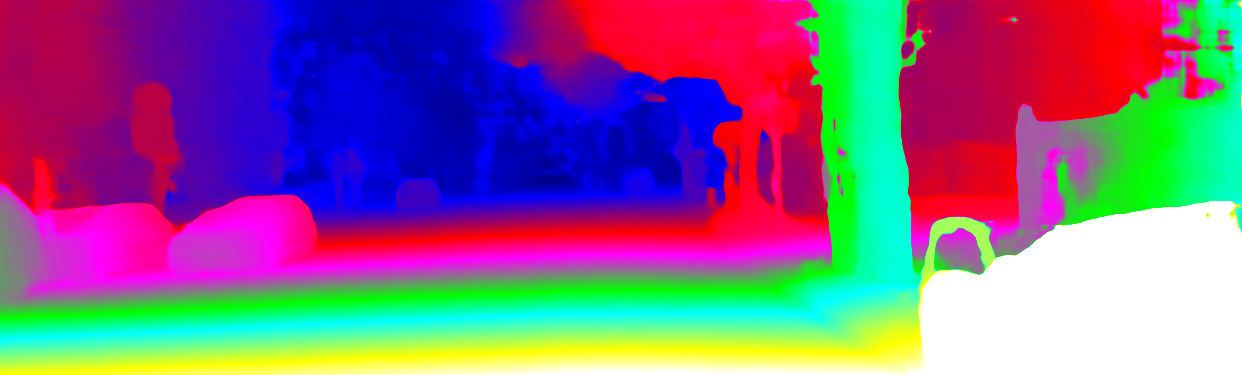}
        \caption{PSMNet (D1-all = 1.43\%) \cite{psmnet}}
    \end{subfigure}
		 \\
     \begin{subfigure}[b]{0.4\textwidth}
        \includegraphics[width=\textwidth]{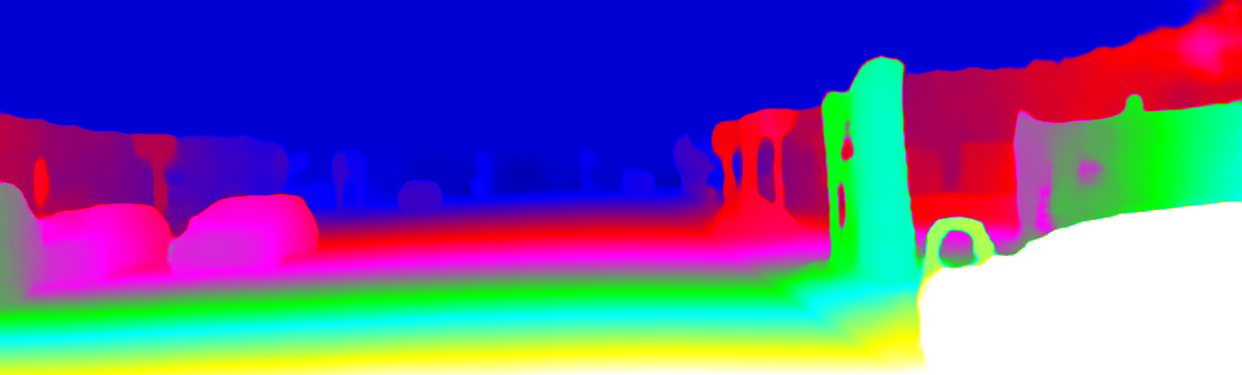}
        \caption{Seg-Stereo (D1-all = 1.12\%) \cite{segstereo}}
    \end{subfigure}
     \qquad
     \begin{subfigure}[b]{0.4\textwidth}
        \includegraphics[width=\textwidth]{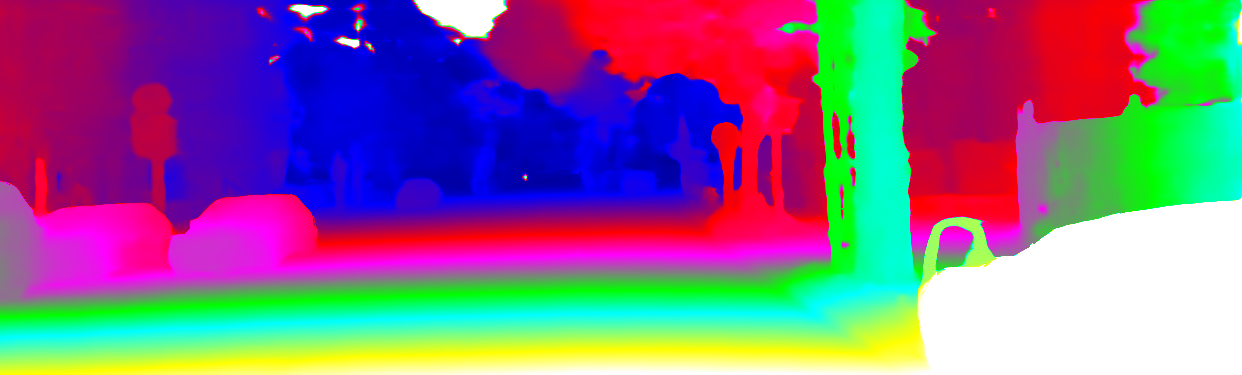}
        \caption{FBA-AMNet-32 (D1-all = 0.67\%) [ours] }
    \end{subfigure}
    \caption{Disparity estimation results of state-of-art methods on Image 4 from KITTI Stereo 2015 test \cite{kitti2015}. The methods are ordered from the least to the most accurate according to D1-all. Our proposed FBA-AMNet is the most accurate, where only $0.67\%$ of the pixels were erroneously estimated, compared to the $1.12\%$ achieved by previous state-of-art methods.}
\label{fig:kitti2015intro}
\end{figure}

In this work, we propose a novel deep neural network architecture for stereo disparity estimation, the atrous multiscale network (AMNet). The proposed network architecture is shown in Fig.~\ref{fig:amnet}. We design our feature extractor by first modifying the standard ResNet-50 backbone to a depthwise separable ResNet (D-ResNet) which makes it feasible to design the network with a larger receptive field without increasing the number of trainable parameters. Second, we propose an atrous multiscale (AM) module, which is designed as a scene understanding module that captures deep global contextual information at multiple scales as well as local details. Our proposed feature extractor constitutes of the D-ResNet followed by the AM module. For cost matching computation, we design a new extended cost volume (ECV) that simultaneously computes different cost matching metrics and constitutes of several cost sub-volumes; a disparity-shifted feature concatenation sub-volume, a disparity-level depthwise correlation sub-volume, a disparity-level feature distance sub-volume. The ECV carries rich information about the matching costs from the different similarity measures. For disparity computation and aggregation, the ECV is processed by a designed stacked AM module which stacks multiple AM modules for multiscale context aggregation. Disparity optimization is done by regression after the soft classification of the quantized disparity bins. 
To enhance the cost volume filtering, and improve the disparity computation and optimization steps, we also propose to learn the foreground-background segmentation as an auxiliary task. The learned foreground background information reinforces disparity estimation similar to image-guided cost-volume filtering. Hence, we train AMNet using multitask learning, in which the main task is disparity estimation and the auxiliary task is foreground-background segmentation. We name the multitask network as foreground-background-aware AMNet (FBA-AMNet). The auxiliary task helps the network have better foreground-background awareness so as to further improve disparity estimation. As discussed above, the optional step of disparity refinement can further improve the estimated disparity.  However, in this work, no refinement has been applied on the estimated disparity. 

\begin{figure*}[t!]
    \centering
    \begin{subfigure}[b]{0.35\textwidth}
        \includegraphics[width=\textwidth]{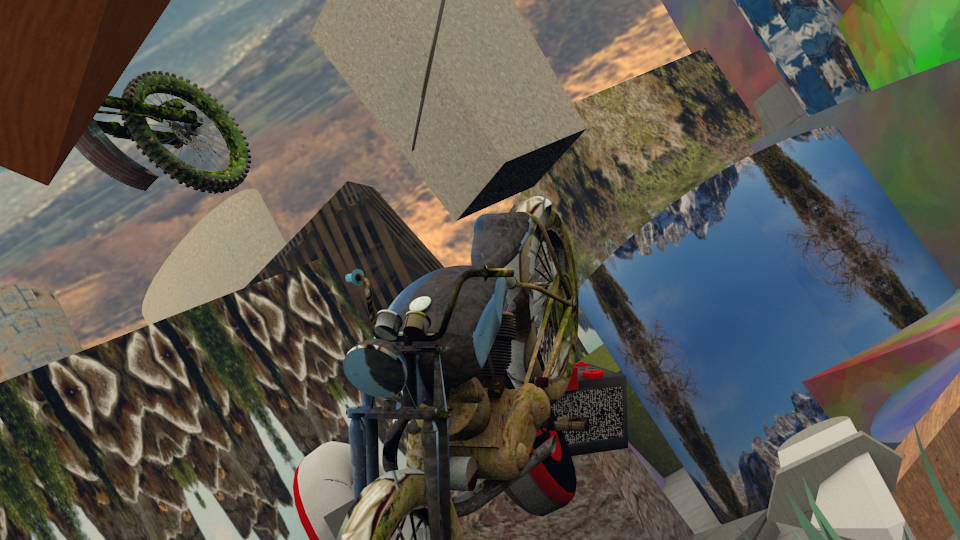}
        \caption{Left input image}
    \end{subfigure}
    \qquad
    \begin{subfigure}[b]{0.35\textwidth}
        \includegraphics[width=\textwidth]{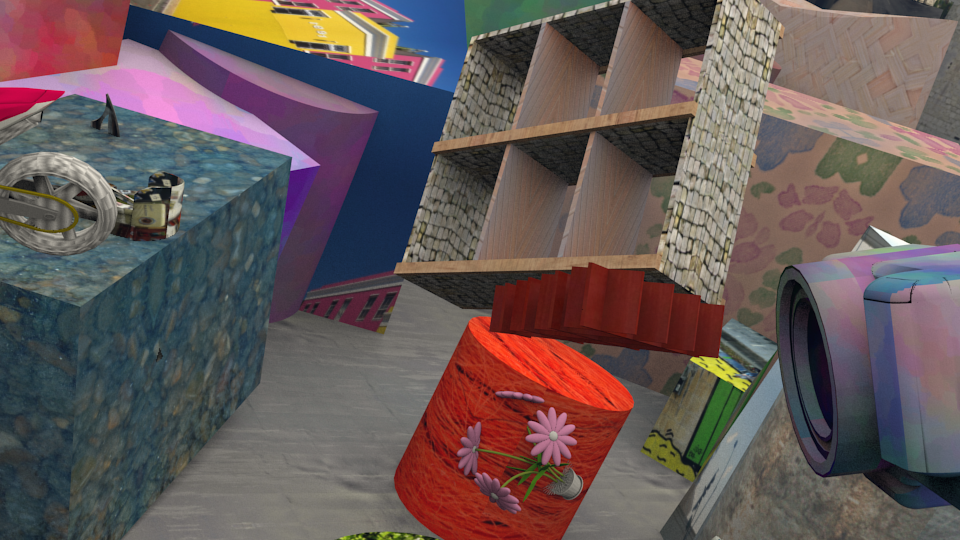}
        \caption{Left input image}
    \end{subfigure}
    
    \begin{subfigure}[b]{0.35\textwidth}
        \includegraphics[width=\textwidth]{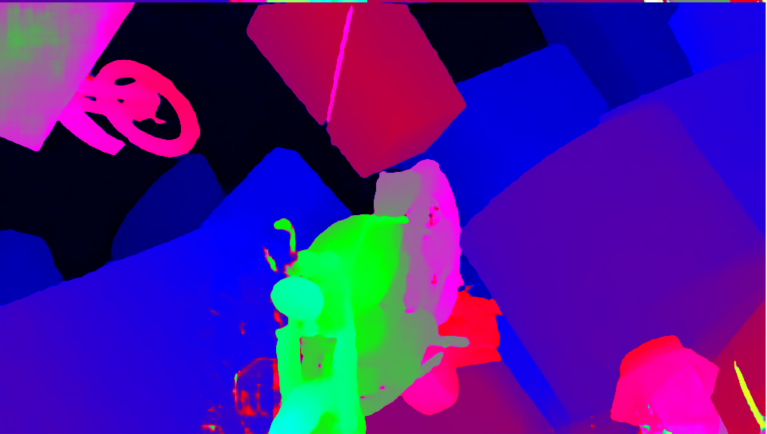}
        \caption{PSMNet \cite{psmnet} (EPE = 2.52)}
    \end{subfigure}
    \qquad
    \begin{subfigure}[b]{0.35\textwidth}
        \includegraphics[width=\textwidth]{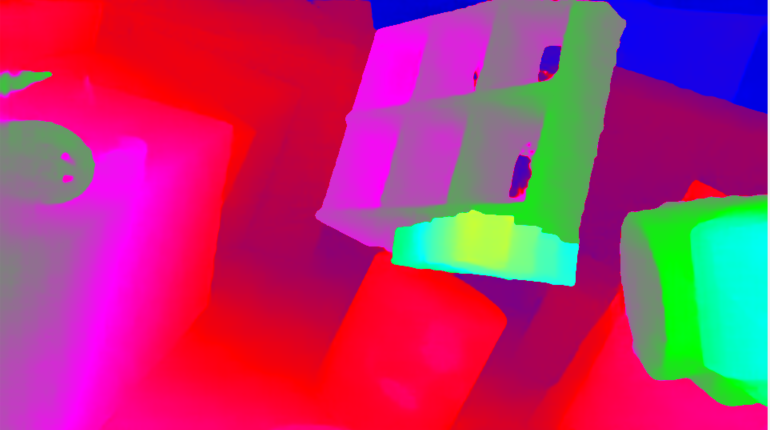}
        \caption{PSMNet \cite{psmnet} (EPE = 1.91)}
    \end{subfigure}
    
    \hspace{2pt}
    \begin{subfigure}[b]{0.35\textwidth}
        \includegraphics[width=\textwidth]{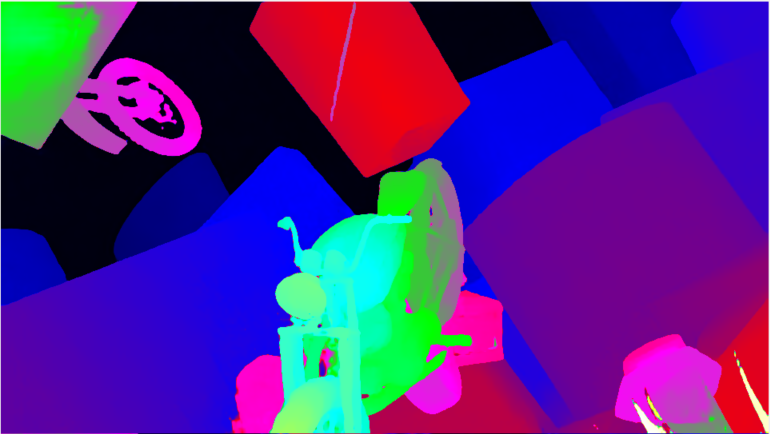}
        \caption{AMNet-32 [this paper] (EPE = 1.14)}
    \end{subfigure}
    \qquad
    \begin{subfigure}[b]{0.35\textwidth}
        \includegraphics[width=\textwidth]{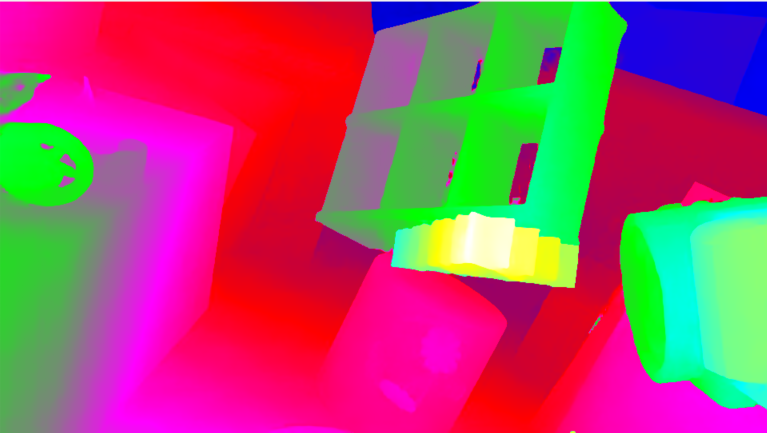}
        \caption{AMNet-32 [\mbox{this paper}] (EPE = 0.61)}
    \end{subfigure}
    
\caption{Visualizations of the disparity estimates of our AMNet-32 model compared to the state- of art method PSMNet \cite{psmnet} for two challenging Sceneflow test images. Disparity estimates by our proposal (AMNet) are the most accurate, where AMNet improved the end point error from $2.52$ to $1.14$ for the left image, and  improved it from $1.91$ to $0.61$ for the right image.}
\label{fig:sceneflow}
\end{figure*}

The proposed AMNets ranked first among all published results on the three most popular disparity estimation benchmarks: KITTI stereo 2015~\cite{kitti2015}, KITTI stereo 2012~\cite{kitti2012}, and Sceneflow~\cite{sceneflow} stereo disparity tests. Some examples showing the superiority  of our proposed atrous multiscale stereo disparity estimation networks are shown in Fig.~\ref{fig:kitti2015intro} and Fig.~\ref{fig:sceneflow}.

The rest of this paper is organized as follows: In the next section, we give more details about previous and related research works. In Sec.~\ref{sec:am}, detailed descriptions of the proposed AMNet are given. Details about FBA-AMNet are given in Sec~\ref{sec:fba}. In Sec.~\ref{sec:exp}, numerical and visualization comparisons with the state-of-art methods on standard benchmark tests are given. Sec.~\ref{sec:conc} concludes this paper.

\section{Related Works}

There has been significant interest to improve the extraction of  contextual information using deep neural networks for better image understanding. The earlier methods used multiscale inputs from an image pyramid~\cite{su1}~\cite{su2}~\cite{su3}~\cite{su5} or implemented probabilistic graphical models~\cite{su6}~\cite{su7}. Recently, models  with spatial pyramid pooling (SPP)~\cite{spp} and encoder-decoder structure have shown great improvements in various computer vision tasks. Zhao et al.~\cite{pspnet} proposed the PSPNet which performs SPP at different grid scales. Chen et al.~\cite{deeplab}~\cite{deeplab2} applied atrous convolutions to the SPP module (ASPP) to process the feature maps using several parallel atrous convolutions with different dilation factors. Newell et al.~\cite{hourglass} designed a stacked hourglass module which stacks an encoder-decoder module three times with shortcut connections to aggregate multiscale contextual information. Chen et al.~\cite{deeplabv3plus2018} further developed the DeepLab v3+ model that combined the ideas of encoder-decoder architecture and ASPP.

Disparity estimation based on a stereo image pair is a well known problem in computer vision. CNN based systems have recently become ubiquitous in solving this problem. In the early work, Zbontar et al.~\cite{MCCNN} proposed a Siamese network to match pairs of image patches for disparity estimation. The network consists of a set of shared convolutional layers, a feature concatenation layer, and a set of fully connected layers for second stage processing and similarity estimation.
 Luo et al.~\cite{luo} developed a faster Siamese network in which cost volume is formed by computing the inner product between the left and the right feature maps and the disparity estimation is forumalated as a  multi-label classification. 

End-to-end neural networks have also been proposed for stereo disparity estimation.  Mayer et al.~\cite{sceneflow}~\cite{flownet}  DispNet which consists of a set of convolution layers for feature extraction, a cost volume formed by feature concatenation or patch-wise correlation, an encoder-decoder structure for second stage processing, and a classification layer for disparity estimation. Motivated by the success of deep neural networks, Kendall et al.~\cite{gcnet} proposed GC-Net. GC-Net uses a deep residual network~\cite{resnet} as the feature extractor, a cost volume formed by disparity-level feature concatenation to incorporates contextual information, a set of $3$D convolutions and $3$D deconvolutions for second stage processing, and a soft argmin operation for disparity regression. To further explore the importance of contextual information, Chang and Chen~\cite{psmnet} proposed the pyramid stereo matching network (PSMNet). Before constructing the cost volume, PSMNet learns the contextual information from the extracted features through a spatial pyramid pooling module. For disparity computation, PSMNet processes the cost volume using a stacked hourglass CNN  which constitutes of three hourglass CNNs. Each hourglass CNN has an encoder-decoder architecture, where the encoder and decoder parts of each hourglass network involve downsampling and upsampling of feature maps, respectively. 

Fusion of semantic segmentation information with other extracted information can result in better scene understanding, and hence has been shown effective in improving the accuracy of challenging computer vision tasks, such as multiscale pedestrian detection~\cite{du2017fused}. Consequently, researchers tried to utilize information from low-level vision tasks such as semantic segmentation or edge detection to reinforce the disparity estimation system.  
 Yang et al.~\cite{segstereo} introduce the SegStereo model, which suggests that appropriate incorporation of semantic cues can rectify disparity estimation. The SegStereo model embeds semantic features to enhance intermediate features and regularize the loss term. Song et al.~\cite{edgestereo} proposed EdgeStereo where edge features are embedded and cooperated by concatenating them to features at different scales of the residual pyramid network, and trained using multiphase training.   

Some works have been dedicated to design disparity refinement networks to improve the depth or disparity estimated from previous state-of-art methods. Fergus et al.~\cite{eigen2014depth}, designed a coarse-to-fine depth refinement module that improved the accuracy of the depth estimated by a single-image depth estimation network. Recently, a refinement module called a convolutional spatial propagation network (CSPN) was proposed, and was trained to refine the output from existing state-of-art networks for single image depth estimation~\cite{eigen2014depth} or stereo disparity estimation~\cite{psmnet}, which improved their accuracies~\cite{CSPN}. A recent work, DispSegNet~\cite{DispSegNet} concatenated semantic segmentation embeddings with the initial disparity estimates before passing them to the second stage refinement network which improved the disparity estimation in ill-posed regions.

\section{Atrous Multiscale Network \label{sec:am}}

In this section, we describe each component of the proposed stereo disparity estimation network. The network architecture of the proposed AMNet is shown in Fig.~\ref{fig:amnet}.

\subsection{Depth Separable ResNet for Feature Extraction}
We propose an efficient feature extractor using depth separable convolutions with residual connections. 
A standard convolution can be decomposed into a depthwise separable convolutions followed by a $1\times 1$ convolution. Depth separable convolutions have recently shown great potential in image classification~\cite{xception}, and has been further developed for other computer vision tasks as a network backbone~\cite{mobilenetv22018}~\cite{deeplabv3plus2018}~\cite{deformable}. Depth separable residual networks have also been proposed for image enhancement tasks such as image denoising~\cite{ren2018dn}.

 Inspired by these works, we design the D-ResNet, as the feature extraction backbone, by replacing standard convolutions with customized depthwise separable convolutions. Our approach differs from previous approaches whose goal is to reduce the complexity. Instead, we use depth separable convolutions to increase the residual network's learning capacity while keeping the number of trainable parameters the same. 
A depthwise separable convolution replaces the three dimensional convolution with two dimensional convolutions done separately on each input channel (depthwise), followed by a pointwise $1 \times 1$ convolution that combines the output of the separate convolutions into an output feature map.  
Let $D_{in}$ and $D_{out}$ represent the number of the input and output feature maps at a convolutional layer, respectively. A standard $3\times 3$ convolutional layer contains $9\times D_{in}\times D_{out}$ parameters, while a depthwise separable convolutional layer contains $D_{in}\times(9+D_{out})$ parameters, which is much smaller for typical choices of $D_{in}$ and $D_{out}$.  

\begin{table}[h!]
\begin{center}
\begin{tabular}{|l|l|l|l|l|l|}
\hline
Index & Type & $D_{out}$ & Str. & Dil. & Repeat \\
\hline
1 & Sepconv & 32 & 2 & 1 & 1\\
2-3 & Sepconv & 32 & 1 & 1 & 2 \\
4-6 & D-ResNet block & 96 & 1 & 1& 3\\
7 & D-ResNet block & 256 & 2 & 1 & 1 \\
8-25 & D-ResNet block &256 & 1 & 1 & 18 \\
26-28 & D-ResNet block & 256 & 1 & 2 & 3\\
\hline
\end{tabular}
\end{center}
\caption{Detailed layer specifications of the D-ResNet. `Repeat' means the current layer or block is repeated a certain number of times. `Str.' and `Dil.' refer to stride and dilation factor.}
\label{tab:dres}
\end{table}

\begin{figure}[h!]
    \centering
    \begin{subfigure}[b]{0.17\textwidth}
        \includegraphics[width=\textwidth]{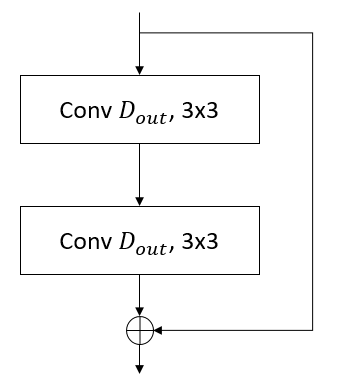}
        \caption{A ResNet block.}
    \end{subfigure}
    ~ 
    \begin{subfigure}[b]{0.29\textwidth}
        \includegraphics[width=\textwidth]{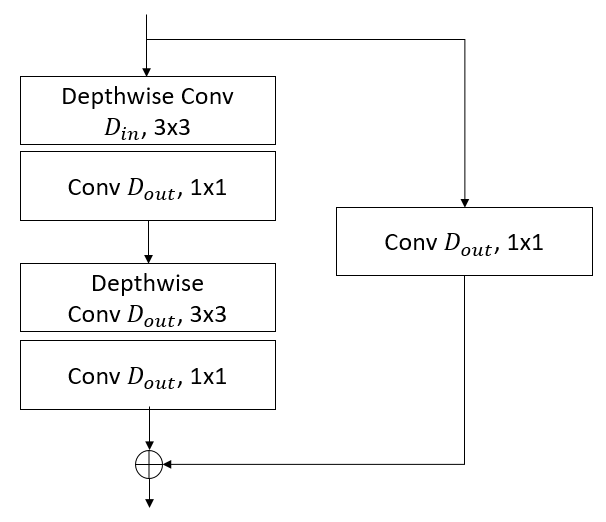}
        \caption{A D-ResNet block.}
    \end{subfigure}
    \caption{Network architecture of the ResNet block (left) and the proposed depth-separable ResNet (D-ResNet) block (right). The D-ResNet block has a pointwise projection layer in the shortcut connection for dimension matching between the input and output features to the block.}
\label{fig:blocks}
\end{figure}

We modified the 50-layer residual network (ResNet) proposed in PSMNet~\cite{psmnet} as a feature extractor, which constitutes of 4 groups of residual blocks, where each residual block consitutes of two convolutional layers with $3 \times 3$ convolutional kernels. The number of residual blocks in the 4 groups are $\{3,16,3,3\}$ respectively. In PSMNet's ResNet, the number of output feature maps are $D_{out} = \{32,64, 128, 128\}$ for the four residual groups, respectively, where $D_{in} = D_{out}$ for all the residual blocks. Since $D_{out}$ is $32$ or larger, a direct replacement of the standard convolutions with a depthwise separable convolution will result in a model with much less number of parameters. However, in our proposed depth-separable ResNet (D-ResNet), we increase $D_{out}$ for the depthwise separable convolutional layers in four residual blocks to be $D_{out} = \{96,256, 256, 256\}$, respectively, where $D_{in}=32$ for the first block, so as to make the number of parameters in the proposed D-ResNet close to that of PSMNet. Thus, the proposed D-ResNet  can learn more deep features than ResNet while having a similar complexity.  Since our proposed depth-separable residual blocks do not necessary have the same number of input and output features, we deploy $D_{out}$ pointwise $1\times1$ projection filters on the shortcut (residual) connection to project the $D_{in}$ input features onto the $D_{out}$ features. 
 Fig.~\ref{fig:blocks} shows a comparison between a standard ResNet block and the proposed D-ResNet block.  ReLU and Batch Normalization are used after each layer. After the D-ResNet backbone, the  widths and heights of the output feature maps are $\frac{1}{4}$th of those of the input image. The network specifications of the D-ResNet backbone are listed in Table~\ref{tab:dres}.

\begin{figure}[t!]
\begin{center}
   \includegraphics[width=1.0\linewidth]{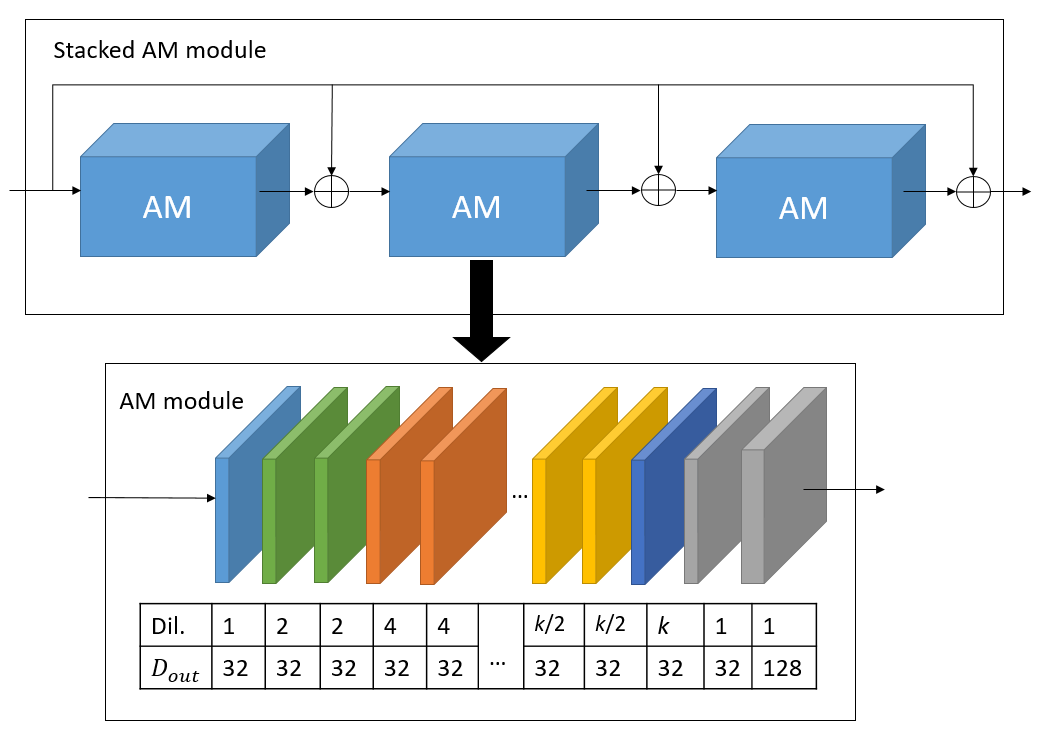}
\end{center}
   \caption{Architecture of the proposed Atrous multiscale (AM) module and the stacked AM (SAM) module. `Dil.' and $D_{out}$ represent the dilation factor and number of output features of each atrous convolutional layer.}
\label{fig:sam}
\end{figure}

\subsection{Atrous Multiscale Context Aggregation }
Since the accuracy of disparity estimation relies on the ability to identify key features at multiple scales, we consider aggregating multiscale contextual information from the deep features. Depth or disparity estimation networks tend to use down-samplings and up-samplings or encoder-decoder architectures, also called hour glass architectures \cite{psmnet,flownet,su2} to aggregate information at multiple scales.  The spatial resolution tends to be lost by pooling or downsampling, 
We use an AM module after the D-ResNet backbone to form the feature extractor. The deep features extracted by the D-ResNet from the stereo image pair are processed by the proposed atrous multiscale (AM) modules before using them to calculate the disparity, as shown in Fig.~\ref{fig:amnet} . Atrous (also called dilated) convolutions provide denser features than earlier methods such as pooling, feature scaling. Inspired by the context network~\cite{dilatedconv} and the hourglass module~\cite{hourglass}~\cite{psmnet}, we design an AM module as a set of $3\times 3$ atrous convolutions with increasing dilation factors such as $[1,2,2,4,4,...,\frac{k}{2},\frac{k}{2},k]$. The dilation factors increase as the AM module goes deeper to increase the receptive field and capture denser multiscale contextual information without losing the spatial resolution. Two $1\times1$ convolutions with dilation factor one are added at the end for feature refinement and feature size adjusting.

\begin{figure*}[t]
\begin{center}
   \includegraphics[width=0.97\linewidth]{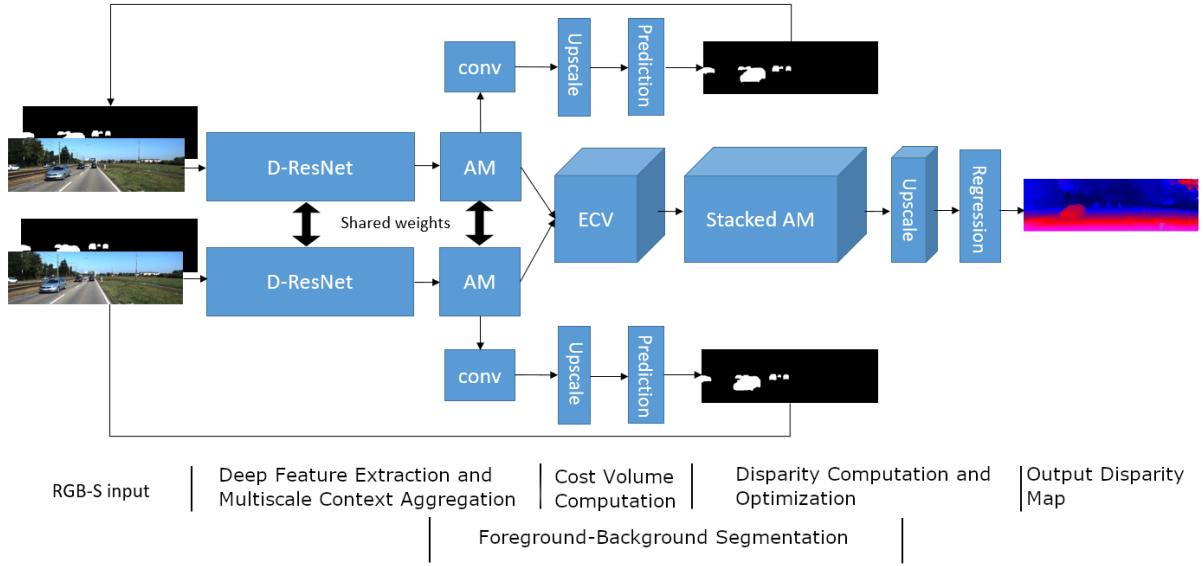}
\end{center}
   \caption{Architecture of proposed multitask Foreground-Background Aware Atrous Multiscale Network (FBA-AMNet) for stereo disparity estimation.}
\label{fig:fbaamnet}
\end{figure*}

\begin{figure*}
    \centering
    \begin{subfigure}[b]{0.3\textwidth}
        \includegraphics[width=\textwidth]{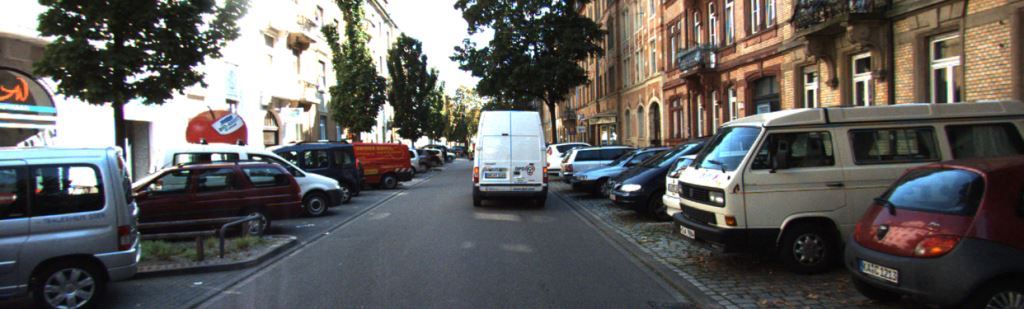}
        \caption{Left input image}
    \end{subfigure}
    \qquad
    \begin{subfigure}[b]{0.3\textwidth}
        \includegraphics[width=\textwidth]{im5.jpg}
        \caption{Left input image}
    \end{subfigure}
		 \\
    \begin{subfigure}[b]{0.3\textwidth}
        \includegraphics[width=\textwidth]{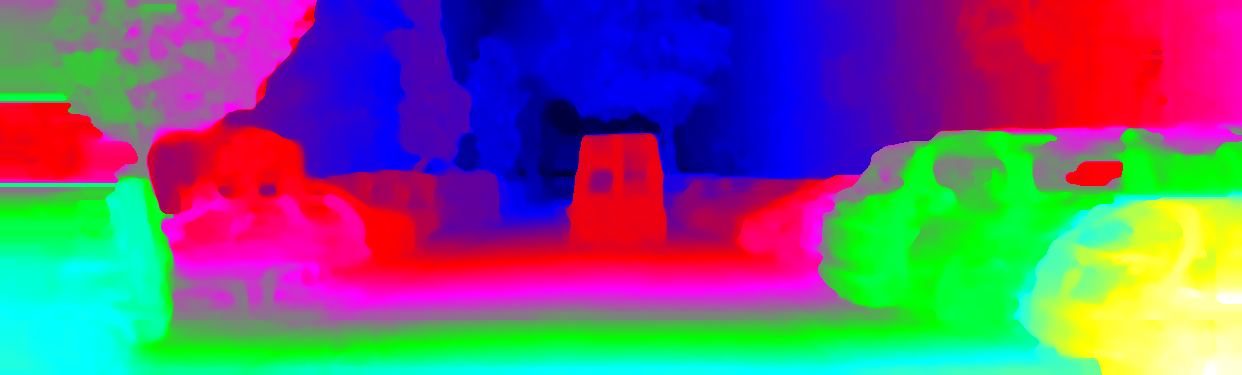}
        \caption{DispSegNet (D1-all = 13.71\%)}
    \end{subfigure}
    \qquad
    \begin{subfigure}[b]{0.3\textwidth}
        \includegraphics[width=\textwidth]{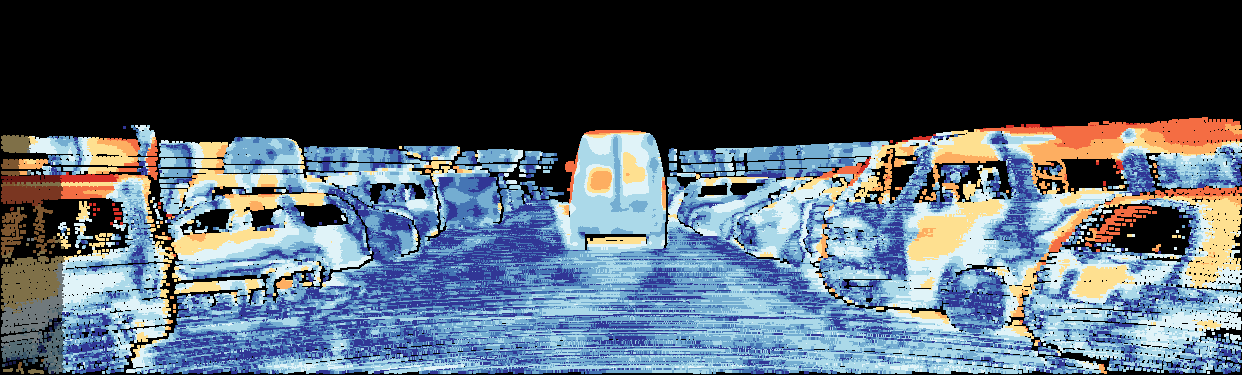}
        \caption{DispSegNet Error \cite{DispSegNet} }
    \end{subfigure}
	  \\
    \begin{subfigure}[b]{0.3\textwidth}
        \includegraphics[width=\textwidth]{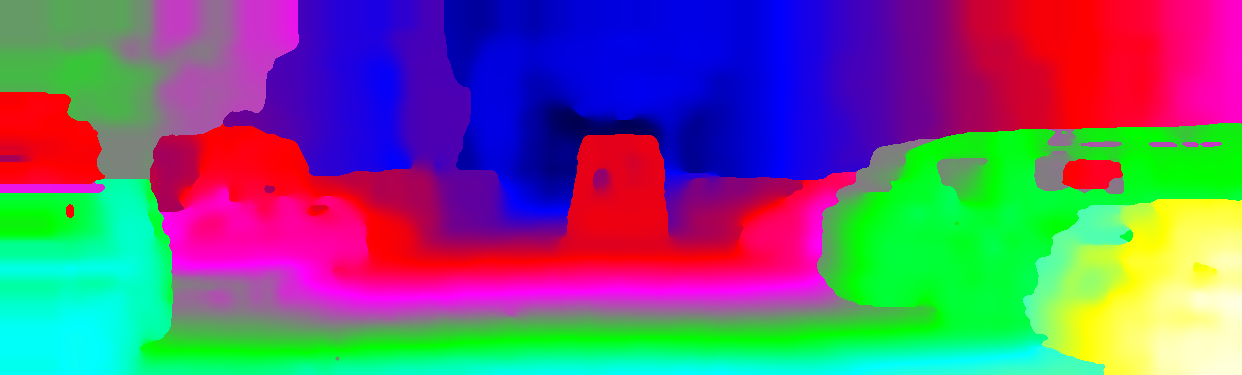}
        \caption{MC-CCN (D1-all = 7.49\%)}
    \end{subfigure}
    \qquad
    \begin{subfigure}[b]{0.3\textwidth}
        \includegraphics[width=\textwidth]{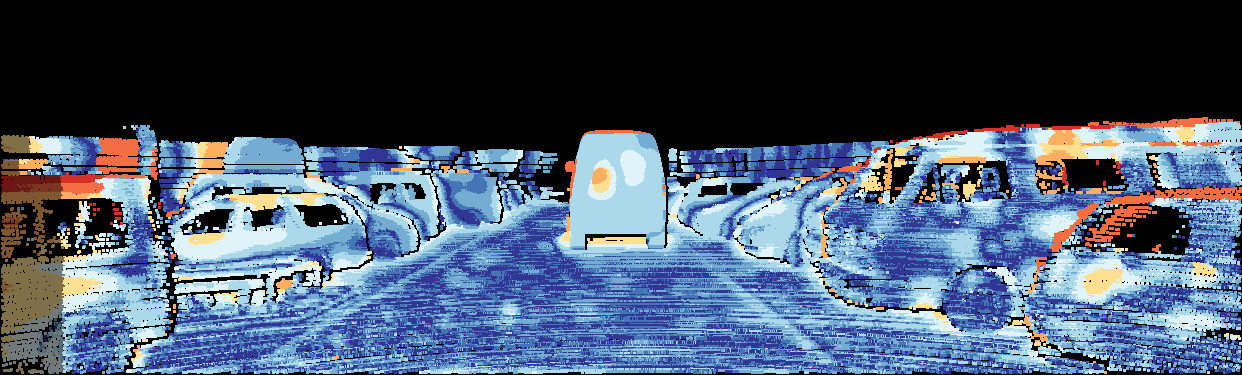}
        \caption{MC-CCN Error \cite{MCCNN} }
    \end{subfigure}
				 \\
    \begin{subfigure}[b]{0.3\textwidth}
        \includegraphics[width=\textwidth]{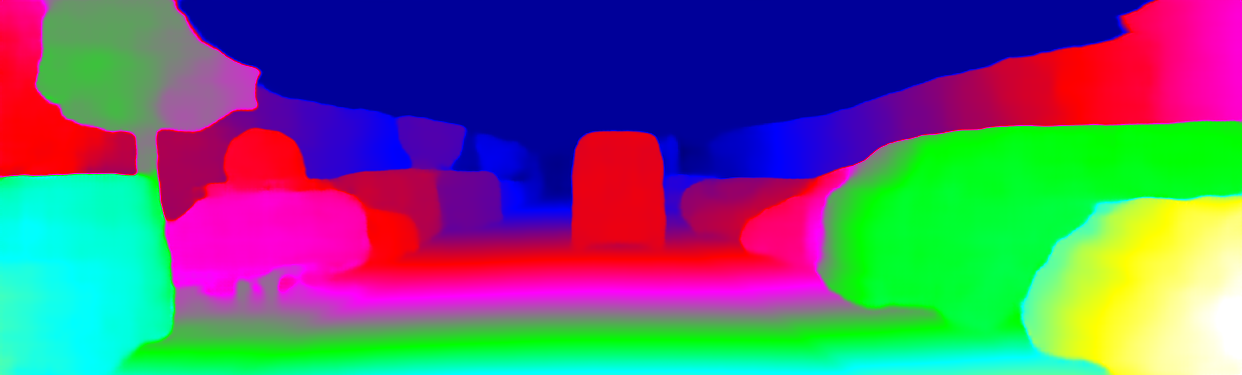}
        \caption{Seg-Stereo (D1-all = 3.52\%)}
    \end{subfigure}
    \qquad
    \begin{subfigure}[b]{0.3\textwidth}
        \includegraphics[width=\textwidth]{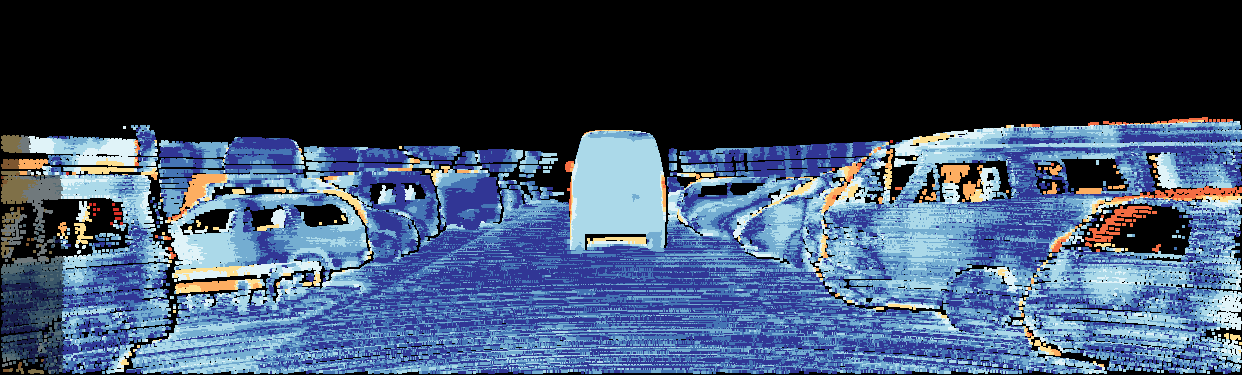}
        \caption{Seg-Stereo Error \cite{segstereo}  }
    \end{subfigure}
		    \\
    \begin{subfigure}[b]{0.3\textwidth}
        \includegraphics[width=\textwidth]{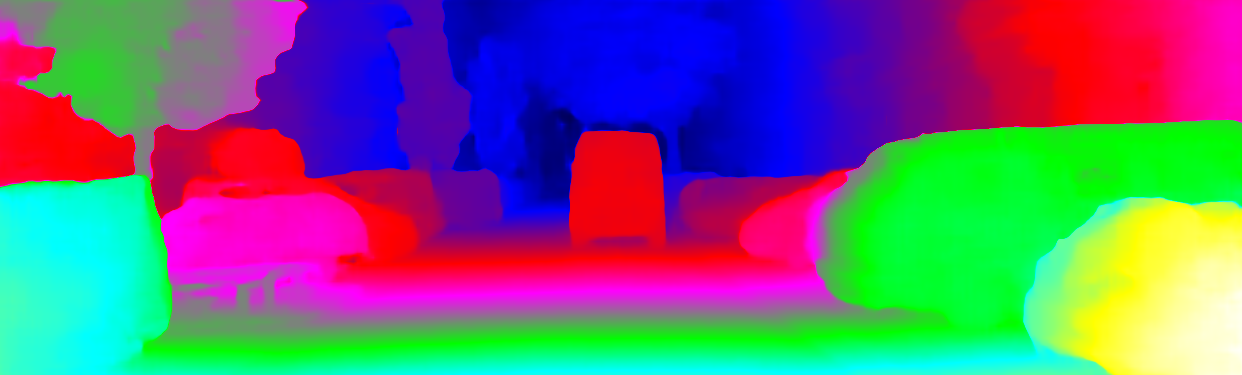}
        \caption{PSMNet (D1-all = 3.45\%)}
    \end{subfigure}
    \qquad
    \begin{subfigure}[b]{0.3\textwidth}
        \includegraphics[width=\textwidth]{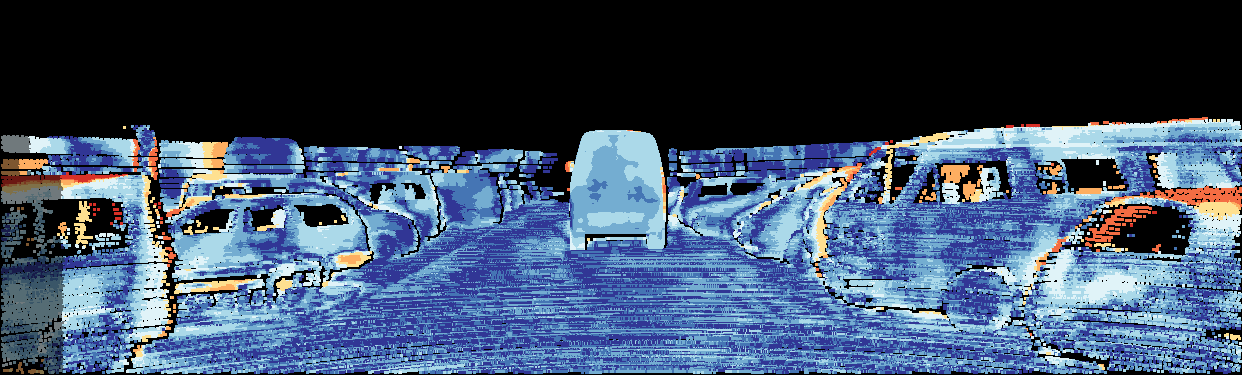}
        \caption{PSMNet Error \cite{psmnet}}
    \end{subfigure}
		\\
    \begin{subfigure}[b]{0.3\textwidth}
        \includegraphics[width=\textwidth]{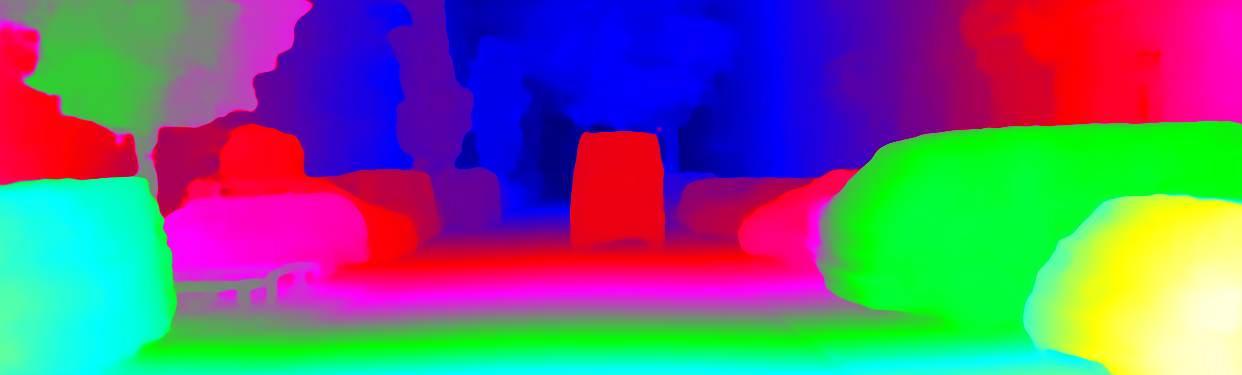}
        \caption{MS-CSPN (D1-all = 2.74\%)}
    \end{subfigure}
    \qquad
    \begin{subfigure}[b]{0.3\textwidth}
        \includegraphics[width=\textwidth]{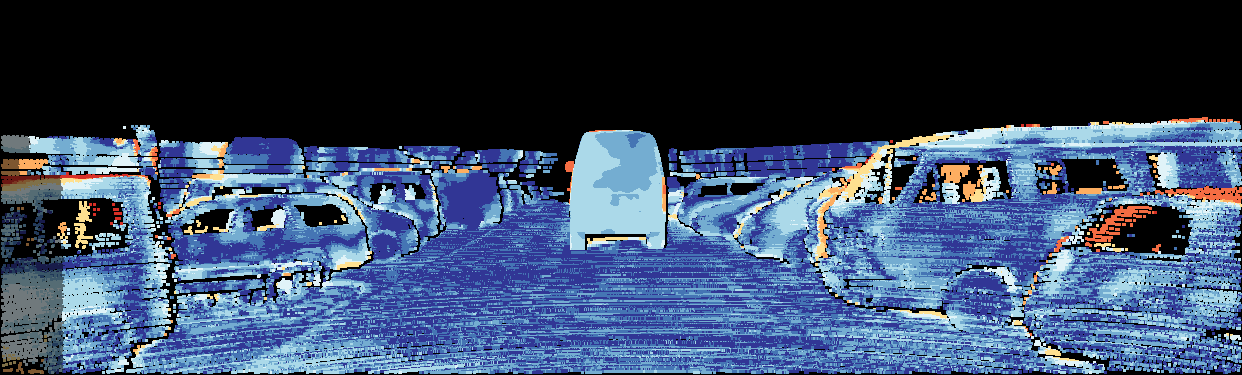}
        \caption{MS-CSPN Error \cite{CSPN}}
    \end{subfigure}
		\\
		 \begin{subfigure}[b]{0.3\textwidth}
        \includegraphics[width=\textwidth]{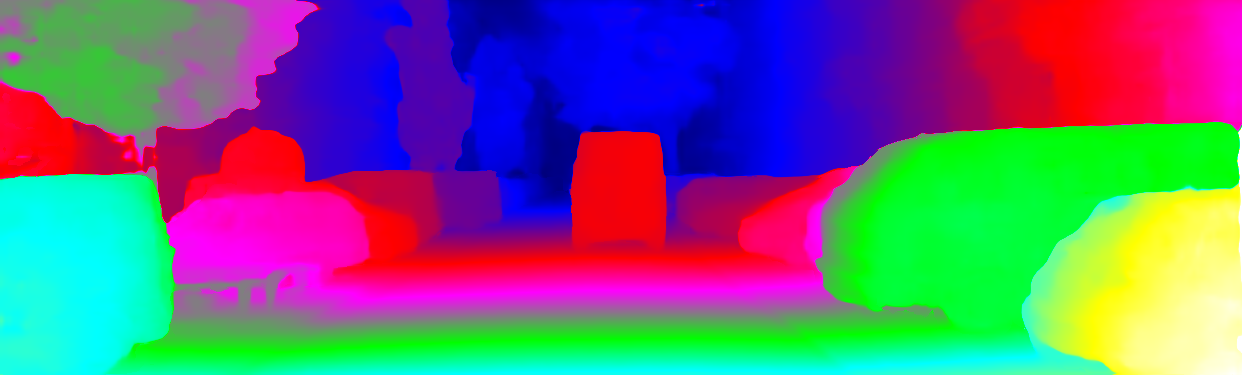}
        \caption{FBA-AMNet-32 (D1-all = 2.34\%)}
    \end{subfigure}
    \qquad
    \begin{subfigure}[b]{0.3\textwidth}
        \includegraphics[width=\textwidth]{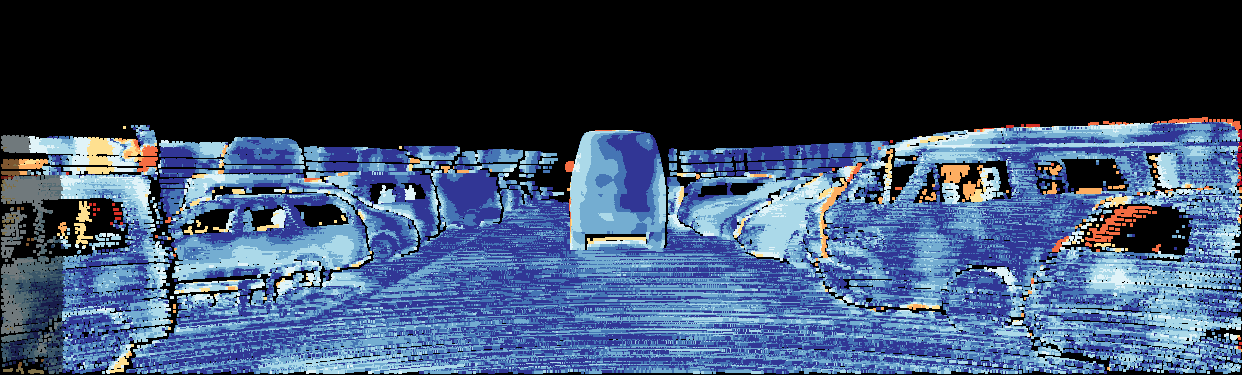}
        \caption{FBA-AMNet-32 Error (this paper) }
    \end{subfigure}
		\caption{Visualizations of the disparity estimates and the errors from the ground truth of our proposed FBA-AMNet are compared to the state-of-art methods when evaluated on Image 5 of KITTI stereo 2015 test set \cite{kitti2015}. Methods are ordered from the least to the most accurate, with FBA-AMNet being the most accurate. }
\label{fig:kitti2015}
\end{figure*}

\subsection{Extended Cost Volume Aggregation \label{sec:ECV}}
We proposed an extended cost volume (ECV) which combines different methods for disparity cost metrics to diversify the information extracted about the true disparity. The cost volume takes as input the deep features extracted by the D-ResNet from the left image and the right image, which are labeled as $F_l$ and $F_r$, respectively. The ECV constitutes of three concatenated cost volumes: disparity-level feature distance, disparity-level depthwise correlation, and disparity-level feature concatenation. Let $D$ be the maximum disparity the AMNet is designed to predict, then let the possible integer disparity shifts be $\mathcal{D}= \{0,1,\ldots, D\}$.  The  three constructed cost volumes, that are concatenated to form the ECV, are described below. 
\begin{enumerate}[label=(\roman*)]
\item \textbf{Disparity-level feature concatenation:} Let $F_r(d)$ refer to the right deep features when  shifted $d$ pixels to the right to align with $F_l$, together with the necessary trimming and zero-padding for $F_l$ to form $F_l(d)$, for $d\in \mathcal{D}$. The left feature maps $F_l(d)$ and the disparity-aligned right feature maps $F_r(d)$ are concatenated for all disparity levels $d\in \mathcal{D}$. Let $W, H, C$, respectively be the width, height, and depth of the feature maps $F_l$ and $F_r$. Then, the size of this cost volume is  $H\times W\times (D+1)\times 2C$.
\item \textbf{Disparity-level feature distance:} The point-wise absolute difference between $F_l$ and $F_r(d)$ is computed at all disparity levels $d$. All the $D+1$ distance maps are packed together to form a sub-volume of size $H\times W\times (D+1)\times C$.
\item \textbf{Disparity-level depthwise correlation:} Following~\cite{flownet}, the correlation between a patch $p_1$ centered at $x_1$ in $F_l$ with a patch $p_2$ centered at $x_2$ in $F_r$ is defined for a square patch of size $2t+1$ as Eq.~\ref{eq:corr}:
\begin{equation} \label{}
    c(x_1,x_2)=\sum_{o \in [-t,t]\times [-t,t]}<F_l(x_1+o),F_r(x_2+o)>.
\label{eq:corr}
\end{equation}
Unlike~\cite{flownet}, instead of computing correlations between $p_1$ with all other patches centered at values within a neighborhood of size $D$ of $x_1$ (expand along the horizontal line), we compute correlations between $p_1$ and its corresponding patches in the aligned $F_r$ across all disparity levels (expand along the disparity level). This results in a sub-volume of size $H\times W\times (D+1)\times 1$. To make the size of the output feature map comparable to other sub-volumes, we implement depthwise correlation. At each disparity level, the depthwise correlations of two aligned patches  are computed and packed across all depth channels for depth indices $i \in \{1,2,\dots,C\}$, by Eq.~\ref{eq:corr2} and Eq.~\ref{eq:corr3}. 
\begin{equation} \label{}
    c^i(x_1,x_1)=\sum_{o \in [-t,t]\times [-t,t]} F_l^i(x_1+o)\times F_r^i(x_1+o),
\label{eq:corr2}
\end{equation}
\begin{equation} \label{}
    c(x_1,x_1)=[c^0(x_1,x_1),c^1(x_1,x_1),...,c^C(x_1,x_1)].
\label{eq:corr3}
\end{equation}   
The depthwise correlation is computed for all patches across all disparity levels, and concatenated to form a cost volume of size $H\times W\times (D+1)\times C$. 
\end{enumerate}

The final ECV has a size of $H\times W\times (D+1)\times 4C$. 
To aggregate the ECV information with more coarse-to-fine contextual information, we propose cascading  three AM modules with shortcut connects within to form the stacked AM module (SAM). The architectures of the proposed AM module and SAM module are shown in Fig.~\ref{fig:sam}. 
Note that due to the introduction of  the disparity dimension by construction of the ECV,  the stacked AM module is implemented with $3$D convolutions to process the ECV.

\subsection{Disparity Optimization}

The smooth $L_1$ loss is used to measure the difference between the predicted disparity $d_i$ and the ground-truth disparity $d_i^{gt}$, at the $i$th pixel. The loss is computed as the average smooth $L_1$ loss over all labeled pixels. During training, three losses are computed separately at the outputs of the three AM modules in the SAM module and summed up to form the final loss, as shown in Eq.~\eqref{eq:loss} and Eq.~\eqref{eq:loss2}:
\begin{equation}
L_{AM}(d_i,d_i^{gt})=\frac{1}{N}\sum_i {L_1}_{\mbox{smooth}} (d_i-d_i^{gt}),
\label{eq:loss}
\end{equation}
\begin{equation}
L=L_{AM_1}+L_{AM_2}+L_{AM_3},
\label{eq:loss2}
\end{equation} 
where $N$ is the total number of labeled pixels. During testing, only the output from the final AM module is used for disparity regression.

At each output layer, the predicted disparity is calculated using the soft argmin operation~\cite{gcnet} for disparity regression. At each pixel, a classification probability is found for each disparity value in $\mathcal{D}$, and the expectation of the $D+1$ disparities is computed as the  disparity prediction, as shown in Eq.~\ref{eq:regression}:

\begin{equation}
d_i=\sum_{j=0}^D j\times p_i^j,
\label{eq:regression}
\end{equation} 
where $p_i^j$ is softmax probability of disparity $j$ at pixel $i$ and $D$ is the maximum disparity value.

\section{Foreground-background Aware Atrous Multiscale Network \label{sec:fba}}
Given the fact that disparities change drastically at the locations where foreground objects appear, we conjecture that a better awareness of foreground objects will lead to a better disparity estimation. In outdoor driving scenes such as KITTI, we define foreground objects as vehicles and humans. In this work, we utilize foreground-background segmentation map to improve disparity estimation. We only differentiate differentiate between foreground and background pixels. 

We considered different methods to utilize the foreground-background segmentation information: The first method is to directly feed the extra foreground-background segmentation information as an additional input besides the RGB image (RGB-S input) to guide the network. This requires accurate segmentation maps in both the training and testing stages. The second method is to train the network as a multitask network. The multitask network is designed to have a shared base and different heads for the two tasks. By optimizing the multitask network towards both tasks, the shared base is trained to have better awareness of foreground objects implicitly, which leads to better disparity estimation. This is the adopted method since it improves the  discrimination capability of the main branch by trying to learn the auxiliary task of FBA, and does not require a standalone segmentation network, which can be quite complex.  
The network structure of the proposed FBA-AMNet is shown in Fig.~\ref{fig:fbaamnet}.
All layers in the feature extractor are shared between the main task of disparity estimation and the auxiliary task of foreground-background segmentation. Beyond the feature extractor, a binary classification layer, an up-sampling layer, and a softmax layer are added for foreground-background segmentation.

The network is trained end-to-end using multitask learning where the loss function is a weighted combination of the losses due to the disparity error and the foreground-background classification error given by $L=L_{disp}+\lambda L_{seg}$, such that $\lambda$ is the relative weight for the segmentation loss.
 We propose an iterative method to train FBA-AMNet. After each epoch, the latest estimated segmentation maps are concatenated with the RGB input to form an RGB-S input to the FBA-AMNet at the next epoch.  
During training, the network keeps refining and utilizing its foreground-background segmentation predictions so as to learn better awareness of foreground objects. At the inference stage, the segmentation task is ignored and we use zero maps as the extra input.

Different from previous works which tried to utilize semantic segmentation \cite{segstereo,DispSegNet}, the proposed foreground-background aware (FBA) network does not differentiate between the different classes of foreground objects or different background classes. We show in our ablation study that this foreground-background awareness gives more accurate disparity estimates than using full semantic segmentation. One reasoning is that foreground-background segmentation can be learned more accurately than full semantic segmentation as it is an easier task to learn, which allows the network optimization to focus more on the main task of disparity estimation.

\section{Experiments \label{sec:exp}}

In this section, we provide numerical and visualization results on public challenges and datasets.

\subsection{Datasets and evaluation metrics}
The proposed method is evaluated on three most popular disparity estimation benchmarks: KITTI stereo 2015~\cite{kitti2015}, KITTI stereo 2012~\cite{kitti2012}, and Sceneflow~\cite{sceneflow}. 

\textbf{KITTI stereo 2015:} The KITTI benchmark provides images in size $376\times 1248$ captured by a pair of stereo camera in real-world driving scenes. KITTI stereo 2015 \cite{kitti2015} consists of 200 training stereo image pairs and 200 test stereo image pairs. Sparse ground-truth disparity maps are provided with the training data. D1-all error is used as the main evaluation metric which computes the percentage of pixels for which the estimation error is $\geq$3px and $\geq$5\% of its ground-truth disparity. 

\textbf{KITTI stereo 2012:} KITTI stereo 2012~\cite{kitti2012} consists of 194 training stereo image pairs and 195 test stereo image pairs. Out-Noc error is used as the main evaluation metric which computes the percentage of pixels for which the estimation error is $\geq$3px for all non-occluded pixels.

\textbf{Sceneflow:} The Sceneflow benchmark ~\cite{sceneflow} is a synthetic dataset suite that contains above 39000 stereo image pairs in size $540\times 960$ rendered from various synthetic sequence. Three subsets contain around 35000 stereo image pairs are used for training (Flyingthings3D training, Monkka, and Driving) and one subset contains around 4000 stereo image pairs is used for testing (Flyingthings3D test). Sceneflow provides complete ground-truth disparity maps for all images. The end-point-error (EPE) is used as the evaluation metric.

\textbf{Middlebury:} The Middlebury stereo benchmark~\cite{middlebury} consists of a training set and a test set with 15 image pairs each in three resolutions, full (F), half (H), and quarter (Q). Ground-truth disparities are provided for the 15 training images. 10 evaluation metrics are used such as the 99-percent error quantile in pixels (A99) and root-mean-square disparity error in pixels (RMS).

\begin{figure*}
    \centering
    \begin{subfigure}[b]{0.45\textwidth}
        \includegraphics[width=\textwidth]{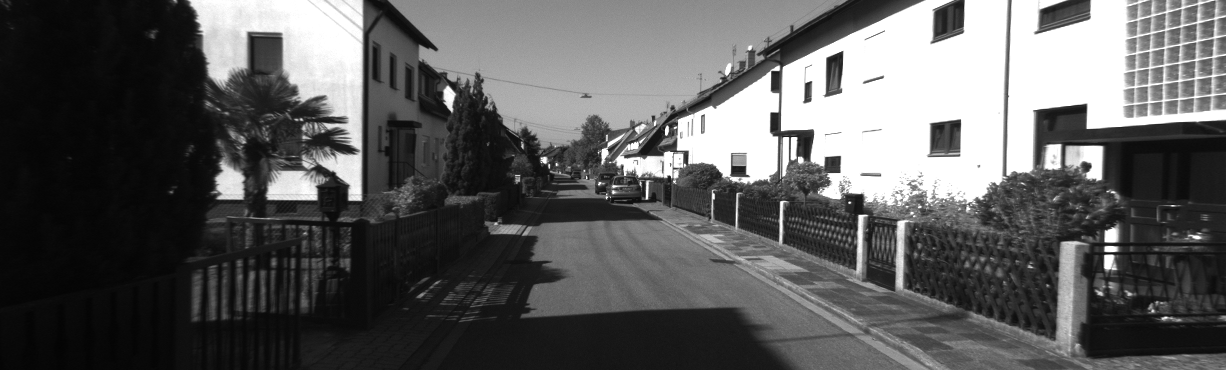}
        \caption{Left input image}
    \end{subfigure}
    \qquad
    \begin{subfigure}[b]{0.45\textwidth}
        \includegraphics[width=\textwidth]{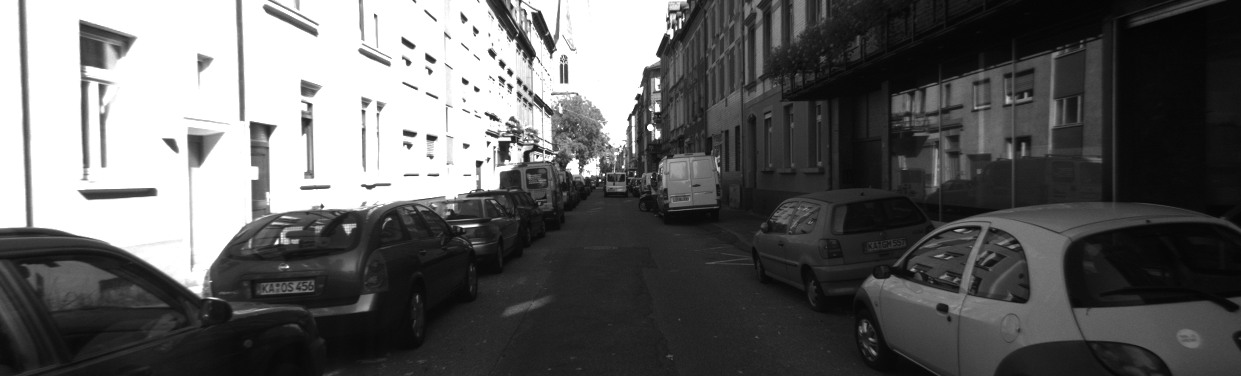}
        \caption{Left input image}
    \end{subfigure}
    
    \begin{subfigure}[b]{0.45\textwidth}
        \includegraphics[width=\textwidth]{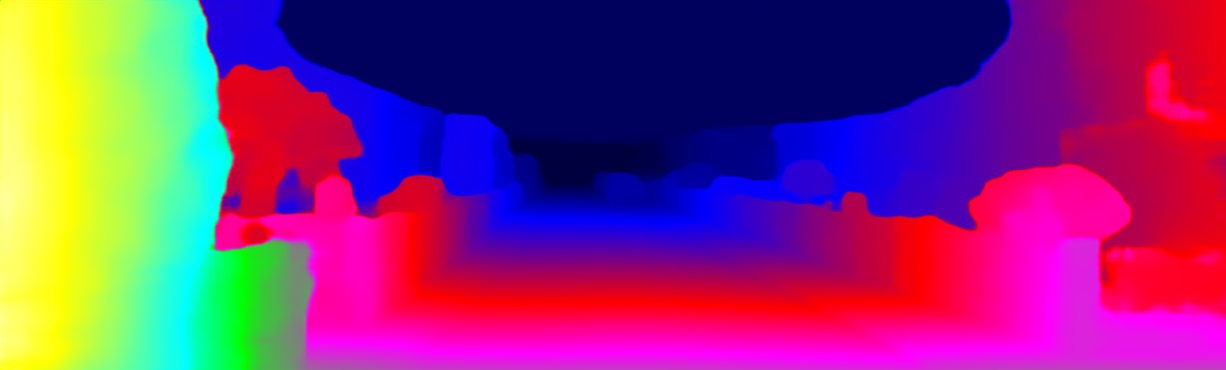}
        \caption{SegStereo (Out-Noc = 3.01\%)}
    \end{subfigure}
    \qquad
    \begin{subfigure}[b]{0.45\textwidth}
        \includegraphics[width=\textwidth]{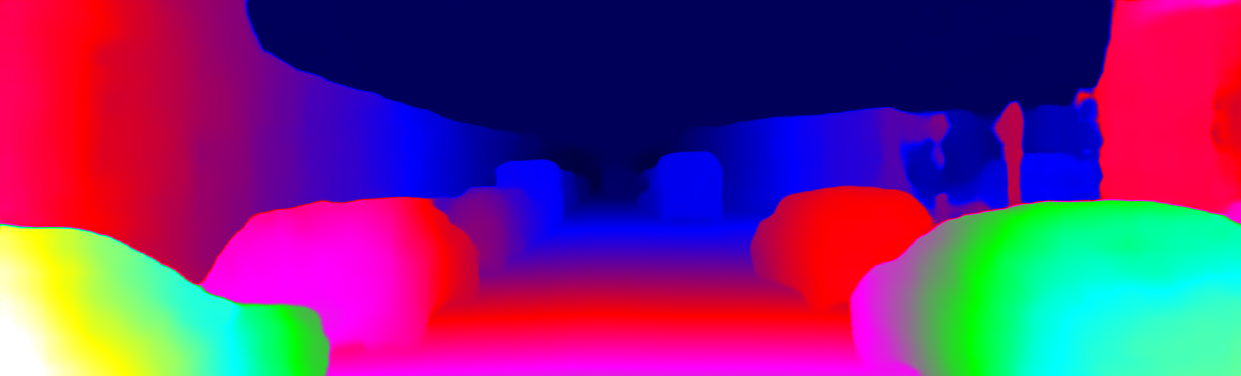}
        \caption{SegStereo (Out-Noc = 1.38\%)}
    \end{subfigure}
    
    \begin{subfigure}[b]{0.45\textwidth}
        \includegraphics[width=\textwidth]{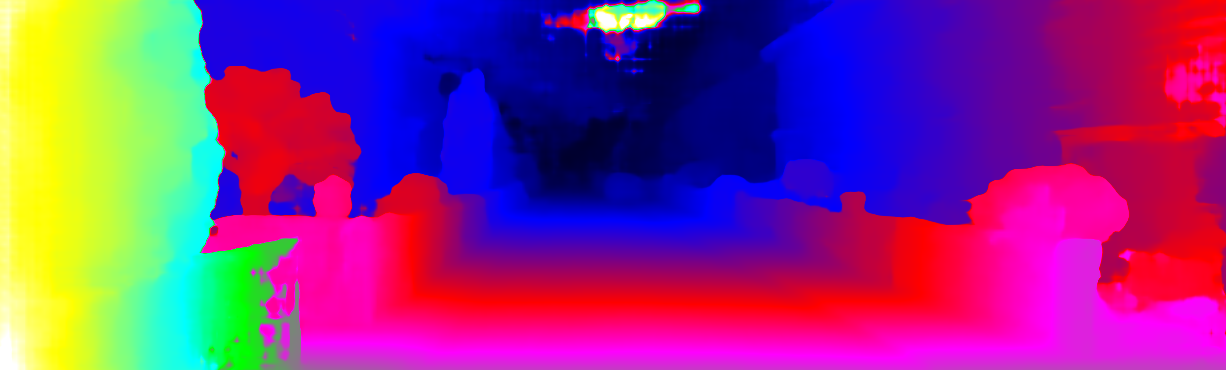}
        \caption{PSMNet (Out-Noc = 2.93\%)}
    \end{subfigure}
    \qquad
    \begin{subfigure}[b]{0.45\textwidth}
        \includegraphics[width=\textwidth]{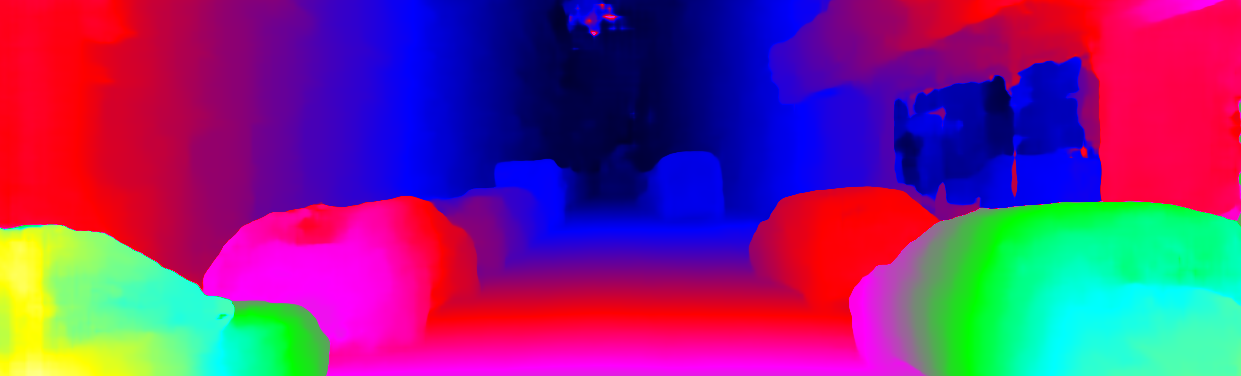}
        \caption{PSMNet (Out-Noc = 1.47\%)}
    \end{subfigure}
    
    \hspace{2pt}
    \begin{subfigure}[b]{0.45\textwidth}
        \includegraphics[width=\textwidth]{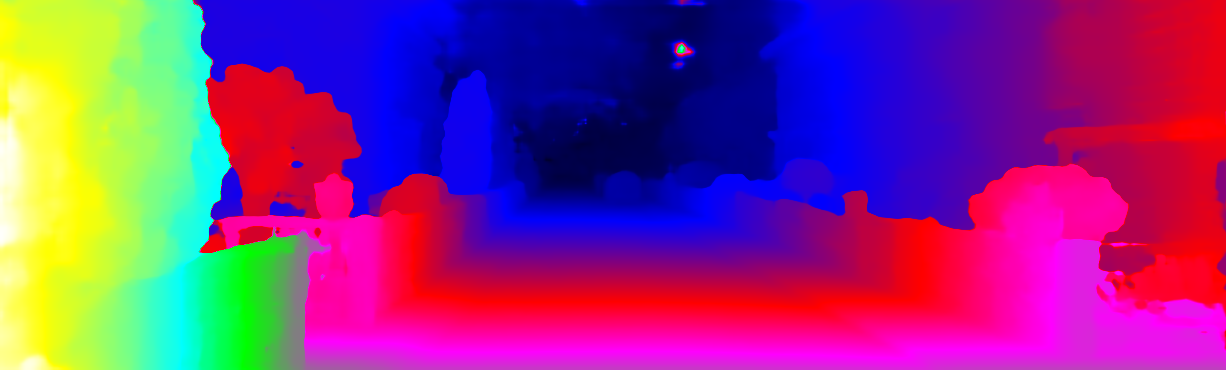}
        \caption{FBA-AMNet-32 (Out-Noc = 1.97\%)}
    \end{subfigure}
    \qquad
    \begin{subfigure}[b]{0.45\textwidth}
        \includegraphics[width=\textwidth]{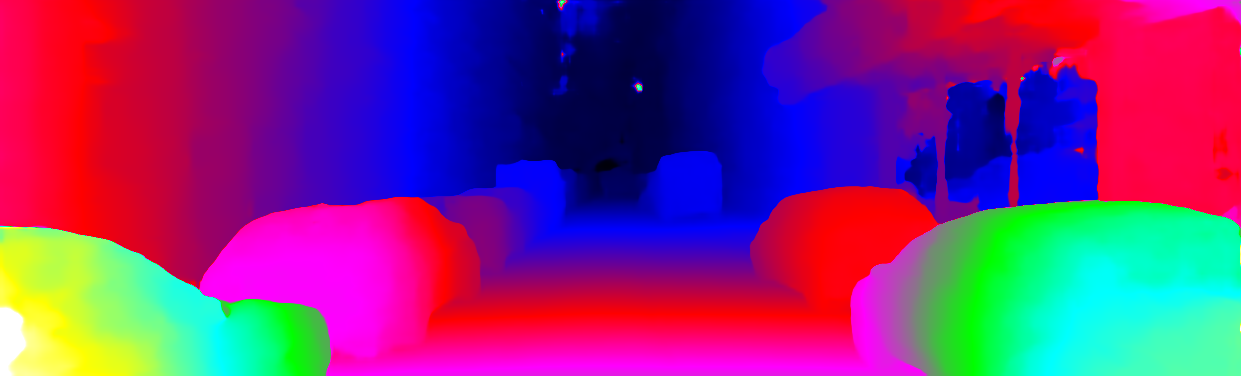}
        \caption{FBA-AMNet-32 (Out-Noc = 1.13\%)}
    \end{subfigure}
    ~ 
    
\caption{Result visualizations of PSMNet, SegStereo, and our FBA-AMNet-32 model for two challenging KITTI stereo 2012 test images \cite{kitti2012} are shown in the left and the right columns. The four rows show the left input images, the results of SegStereo, the results of PSMNet, and our results, respectively. Out-Noc error is shown with each visualization.}
\label{fig:kitti2012}
\end{figure*}

\begin{table*}[t!]
\begin{center}
\begin{tabular}{|l|l|l|l|l|l|l|l|}
\hline
 & \multicolumn{3}{|c|}{All pixels} & \multicolumn{3}{|c|}{Non-Occluded pixels}& \\
\hline
Method & D1-bg & D1-fg & \textbf{D1-all} & D1-bg & D1-fg & D1-all & Runtime\\
\hline\hline
GC-Net~\cite{gcnet} & 2.21\%&	6.16\%&	2.87\% & 2.02\% &5.58\%&2.61\% & 0.9 s\\
PDSNet~\cite{pds} & 2.29\%	&4.05\%&	2.58\% & 2.09\%	&3.68\%	&2.36\%&0.5 s\\
PSMNet~\cite{psmnet} & 1.86\% & 4.62\% & 2.32\%& 1.71\%	&4.31\%	&2.14\%	&0.41 s\\ 
SegStereo~\cite{segstereo} & 1.88\%&	4.07\%&	2.25\% &1.72\%&3.41\%&	2.00\%&0.7 s\\
EdgeStereo~\cite{edgestereo} & 1.87\%&3.61\%& 2.16\% & 2.12\%	&3.85\%	&2.40\%&0.27 s\\
MC-CSPN~\cite{CSPN} & 1.56\%&3.78\%& 1.93\% & 2.12\%	&3.85\%	&2.40\%&0.9 s\\
\hline
AMNet-8 & {1.64\%} & {3.96\%} & {2.03\%} & {1.50\%} &	{3.75}\% & 	{1.87\%} & 0.7 s \\
AMNet-32 & {1.60\%} & {3.81\%} & {1.97\%} & {1.43\%} & {3.48\%} & {1.77\%} & 0.9 s\\
FBA-AMNet-8 & {1.60\%} & {3.88\%} & {1.98\%} & {1.45\%}	& {3.74}\% &\ {1.82\%} &0.7 s\\
FBA-AMNet-32&  \textbf{1.53\%} &	\textbf{3.43\%}& \textbf{1.84\%} & \textbf{1.39\%}	& \textbf{3.20\%} & \textbf{1.69\%} & 0.9 s\\	
\hline

\end{tabular}
\end{center}
\caption{Performance comparisons of our AMNet and FBA-AMNet models with the top published methods on the KITTI stereo 2015 test set \cite{kitti2015}. D1-bg, D1-fg, D1-all refer to disparity error evaluation on the static background pixels, the dynamic foreground pixels, and on all pixels, respectively.}
\label{tab:kitti2015}
\end{table*}

\subsection{Implementation details}
We first train an AMNet-8 and an AMnet-32 from scratch on the Sceneflow training set~\cite{sceneflow}. For the two models, the dilation factors of the atrous convolutional layers in the AM module are set to $[1,2,2,4,4,8,1,1]$ and $[1,2,2,4,4,8,8,16,16,32,1,1]$, respectively. The maximum disparity $D$ is set to $192$. The parameter $t$ in the ECV is set to $0$. The weight $\lambda$ for the segmentation loss is set to $0.5$. For a pair of input images, two patches in size $256\times 512$ at a same random location are cropped as inputs to the network. All pixels with a ground-truth disparity larger than $D$ are excluded from training. The model is trained end-to-end with a batch size of $16$ for $15$ epochs with the Adam optimizer. The learning rate is set to $10^{-3}$ initially and is decreased to $10^{-4}$ after $10$ epochs. 
All the models are implemented with PyTorch and trained on NVIDIA GPUs.

We fine-tune four models: an AMNet-8, an AMNet-32, a FBA-AMNet-8, and a FBA-AMNet-32 on KITTI from our pre-trained Sceneflow AMNet-8 and AMNet-32 models. The FBA-AMNet models are trained using the iterative training method described in Sec.~\ref{sec:fba} with a batch size of $12$ for $1000$ epochs with the Adam optimizer. The learning rate is set to $10^{-3}$ initially and is decreased to $10^{-4}$ after $600$ epochs. We increase the learning rate to $10$ times larger for the new layers. Other settings are the same when training on the Sceneflow test set~\cite{sceneflow}. The foreground-background segmentation maps are initialized as zeros for the first epoch.

We only trained FBA-AMNet on the KITTI benchmark datasets. Due to the fact that segmentation labels in the Sceneflow test set~\cite{sceneflow} are not consistent across scenes or objects, and they are lacking in the Middlebury set, we don't train the FBA-AMNet on the Sceneflow or on the Middlebury datasets, where we only trained AMNet. For the Middlebury benchmark~\cite{middlebury}, we fine-tune our pre-trained AMNet-32 model on the $15$ Middlebury training images in quarter resolution, using the same experiment settings as with the KITTI AMNet-32 model.

\begin{table*}[t!]
\begin{center}
\begin{tabular}{|l|l|l|l|l|}
\hline
Method & \textbf{Out-Noc} & Out-Occ & Ave-Noc & Ave-Occ\\
\hline\hline
PDSNet~\cite{pds} & 1.92\%&	2.53\%	&0.9 px&	1.0 px	\\
GC-Net~\cite{gcnet} & 1.77\%&	2.30\%&	0.6 px&	0.7 px\\
EdgeStereo~\cite{edgestereo} & 1.73\%	&2.18\%	&0.5 px	&0.6 px \\
SegStereo~\cite{segstereo} & 1.68\%&	2.03\%&	0.5 px& 0.6 px\\
PSMNet~\cite{psmnet} & 1.49\% &	1.89\%&	0.5 px&	0.6 px\\
\hline
AMNet-8  & {1.38}\%&	{1.79}\%&	0.5 px&	{0.5 px} \\
AMNet-32  & {1.33}\%&	{1.74}\%&	0.5 px&	{0.5 px} \\
FBA-AMNet-8  & {1.36}\%&	{1.76}\%&	0.5 px&	{0.5 px} \\
\textbf{FBA-AMNet-32}  & \textbf{1.32\%}&	\textbf{1.73}\%&	\textbf{0.5} px&	\textbf{0.5 px} \\
\hline

\end{tabular}
\end{center}
\caption{Performance comparisons on the KITTI stereo 2012 test set \cite{kitti2012}. The error threshold is set to $3$.}
\label{tab:kitti2012}
\end{table*}

\begin{table*}[t!]
\begin{center}
\begin{tabular}{|l|l|l|l|l|l|l|}
\hline
Method & GC-Net~\cite{gcnet} & DispNetC~\cite{sceneflow} & PSMNet~\cite{psmnet} & AMNet-8 & AMNet-32\\
\hline
EPE & 2.51 & 1.68 & 1.09 & \textbf{0.81} & \textbf{0.74} \\
\hline
\end{tabular}
\end{center}
\caption{Performance comparisons on the Sceneflow test set. All results are reported in EPE.}
\label{tab:sceneflow}
\end{table*}

\subsection{Experimental results}
\textbf{Performance on the KITTI stereo 2015 test set:} We submitted our estimated disparity maps to the KITTI server to evaluate our four models, AMNet-8, AMNet-32, FBA-AMNet-8, and FBA-AMNet-32, on the KITTI stereo 2015 test set \cite{kitti2015} and compare it with all published methods on all evaluation settings. The results are shown in Table~\ref{tab:kitti2015}. All our four models perform better than published state-of-art methods on D1-all with significant margins. The FBA-AMNet-32 model lowers the D1-all error on all pixels to $1.84\%$, compared to EdgeStereo which is the previous best result with an end-to-end network whose disparity maps have $17.4\%$ more errors then FBA-AMNet-32. Our end-to-end FBA-AMNet is also better than two stage solutions like MC-CSPN~\cite{CSPN} which added a depth refinement head on top of PSMNet~\cite{psmnet} to improve its performance. 
Visualization of the disparity maps and comparisons with the state-of-art methods on two challenging scenes from the KITTI test set can be observed in Fig.~\ref{fig:kitti2015intro} and Fig.~\ref{fig:kitti2015}. 
The D1-all error for all pixels is computed for each method, and demonstrates that the proposed FBA-AMNet has the least percentage of pixels with erroneous disparity estimates.

\textbf{Performance on the KITTI stereo 2012 test set:} Performance comparisons on the KITTI stereo 2012 test set \cite{kitti2012} are shown in Table~\ref{tab:kitti2012}. Being consistent with KITTI stereo 2015, our four models significantly outperform all other published methods on all evaluation settings. The FBA-AMNet-32 model decreases the Out-Noc to $1.32\%$, with a relative gain of $11.4\%$ compared to the previous best result reported at $1.49\%$. Note that only results with an error threshold of $3$ are reported here, and are consistent with the results for other error thresholds as well. Disparity map visualizations with FBA-AMNet-32, PSMNet, and SegStereo on two challenging KITTI stereo 2012 test images are shown in Fig.~\ref{fig:kitti2012}. The Out-Noc error is computed for each method and confirms the superiority of FBA-AMNet.

\textbf{Performance on the Sceneflow test set:} We compare the AMNet-8 model and the AMNet-32 model with all published methods on the Sceneflow test set~\cite{sceneflow}. Both of our models outperform other methods with large margins. Results reported in EPE are shown in Table~\ref{tab:sceneflow}. Our AMNet-32 model pushes EPE to $0.74$, with a relative gain of $32.1\%$ compared to the previous best result at $1.09$. 
Visualizations of the disparity maps generated by AMNet-32 and PSMNet on two Sceneflow test images are shown in Fig.~\ref{fig:sceneflow}, where the EPE is computed for each method.

\textbf{Performance on the Middlebury test set:} Performance comparisons on Middlebury~\cite{middlebury} are shown in Table~\ref{tab:Middlebury}. AMNet-32 achieves 106 on the A99 test dense metric, which ranks first among all submissions using quarter resolution images, and fourth among all published submissions.

\begin{table}[h!]
\begin{center}
\begin{tabular}{|l|l|l|l|l|l|l|}
\hline
Method & iResNet & DN-CSS & NOSS & AMNet-32 & PSMNet\\
\hline
Resolution & H & H & H & Q & Q \\
\hline
A99-dense-all & 67.8 & 82.0 & 104 & \textbf{106} & 106 \\
rms-dense-all & 13.9 & 16.8 & 19.8 & \textbf{22.9} & 23.3 \\
\hline
\end{tabular}
\end{center}
\caption{Performance comparisons on the Middlebury test set~\cite{middlebury}. All results are reported in EPE.}
\label{tab:Middlebury}
\end{table}

\subsection{Ablation Study}
In this subsection, we analyze the effectiveness of each component of the proposed architecture in details. We conduct most of the analysis on the Sceneflow test set~\cite{sceneflow}, since KITTI only allows a limited number of evaluations  on the test set.

\subsubsection{AMNet versus FBA-AMNet on foreground pixels}
Compared to the AMNet, the FBA-AMNet is designed and trained to generate smoother and more accurate shapes for foreground objects, which leads to finer disparity maps. We visualize the disparity estimation results of the AMNet-32 model and the FBA-AMNet-32 model on two challenging foreground objects from KITTI test images in Fig.~\ref{fig:fgcomp}. The visualizations support the fact that the FBA-AMNet is able to generate finer boundary details for the foreground objects.

\subsubsection{D-ResNet versus ResNet-50 as network backbone}
We explore how the modifications to the network backbone from a ResNet-50 to our proposed D-ResNet change performance and complexity. We compare three models: the AMNet-32 model using PSMNet's ResNet-50~\cite{psmnet} as the network backbone, the AMNet-32 model after modifying the ResNet-50 by directly replacing the standard convolutions with depthwise separable convolutions, and our proposed D-ResNet specified in Table~\ref{tab:dres}. The results on the Sceneflow test set~\cite{sceneflow} and the number of parameters of each model are shown in Table~\ref{tab:backbone}. We can see that D-ResNet performs better then the reference ResNet-50 and with less parameters.

\begin{table}[h!]
\begin{center}
\begin{tabular}{|l|l|l|}
\hline
Backbone & EPE & \# parameters\\
\hline
ResNet-50 & 0.79& 4.81 million \\
ResNet-50 (sep conv) & 0.81 & 1.72 million \\
D-ResNet & \textbf{0.74} & 4.37 million \\
\hline
\end{tabular}
\end{center}
\caption{Performance and complexity comparisons of three models using different network backbones for feature extraction. Results are reported on the Sceneflow test set~\cite{sceneflow}.}
\label{tab:backbone}
\end{table}

\subsubsection{Ablation study for the extended cost volume}
We perform an ablation study for the extended  with seven models modified from the AMNet-32 model by using different combinations of the three constituent volumes of the ECV introduced in Sec.~\ref{sec:ECV}. Comparisons of the results on the Sceneflow test~\cite{sceneflow} set are shown in Table~\ref{tab:ecv}. The results show that the disparity-level feature distance volume is more effective than the other two, and a combination of the three volumes to form the ECV leads to the best performance.

\begin{table}[h!]
\begin{center}
\begin{tabular}{|l|l|l|}
\hline
Cost volume  & EPE & Feature size\\
\hline
Dist. & 0.82 & H$\times$W$\times$(D+1)$\times$C\\
Corr. & 0.85 & H$\times$W$\times$(D+1)$\times$C\\
FC & 0.84 & H$\times$W$\times$(D+1)$\times$2C\\
Dist. + Corr. & 0.78 & H$\times$W$\times$(D+1)$\times$2C\\
Dist. + FC & 0.76 &  H$\times$W$\times$(D+1)$\times$3C\\
Corr. + FC & 0.8 & H$\times$W$\times$(D+1)$\times$3C\\
ECV & \textbf{0.74} & H$\times$W$\times$(D+1)$\times$4C\\
\hline
\end{tabular}
\end{center}
\caption{Performance and feature size comparisons of models using different cost volumes. `Dist.', `Corr.', and `FC' refer to the disparity-level feature distance, the disparity-level depthwise correlation, and the disparity-level feature concatenation, respectively. All results are reported on the Sceneflow test set~\cite{sceneflow} in EPE.}
\label{tab:ecv}
\end{table}

\subsubsection{Going deeper with AM module}
 Table~\ref{tab:deeper} shows how different network architectures of the AM module affect the performance and speed of the AMNet model by setting its maximum dilation factor $k$ to $4$, $8$, $16$, and $32$. We confirm that a deeper structure allows the AM module to aggregate more multiscale contextual information and leads to a finer feature representation and more accurate disparity estimation, at the expense of extra computational cost.

\begin{table}[h!]
\begin{center}
\begin{tabular}{|l|l|l|}
\hline
k & EPE & Runtime\\
\hline
4 & 0.86 & 0.6 s\\
8 & 0.81 & 0.8 s\\
16 & 0.77 & 0.9 s\\ 
32 & \textbf{0.74} & 1.1 s\\
\hline
\end{tabular}
\end{center}
\caption{Performance and run time per image comparisons at different AM module depths defined by the maximum dilation factor $k$. All results are reported on the Sceneflow test set~\cite{sceneflow}. Test images are in size $540\times 960$.}
\label{tab:deeper}
\end{table}

\subsubsection{Performance visualizations of the foreground-background segmentation task}
\begin{figure}
    \centering
    \begin{subfigure}[b]{0.4\textwidth}
        \includegraphics[width=\textwidth]{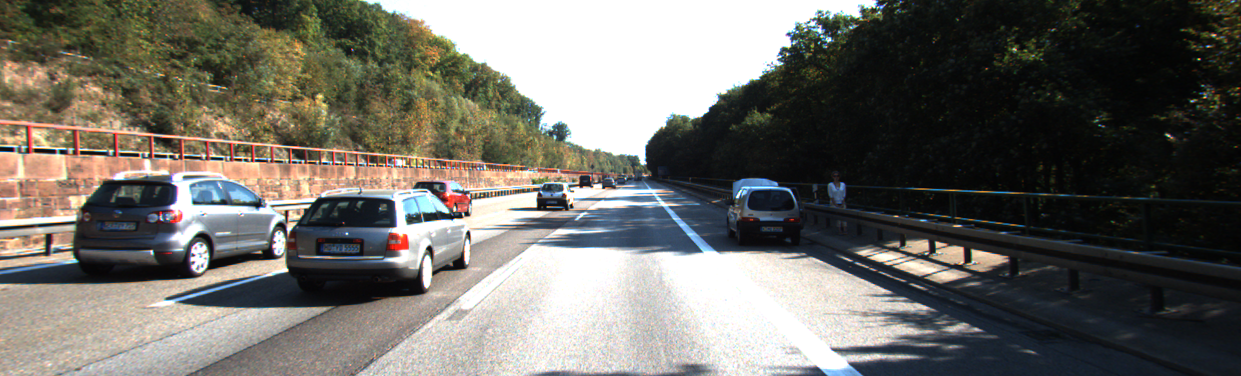}
        \caption{The input image.}
    \end{subfigure}
		\\
    ~ 
    \begin{subfigure}[b]{0.4\textwidth}
        \includegraphics[width=\textwidth]{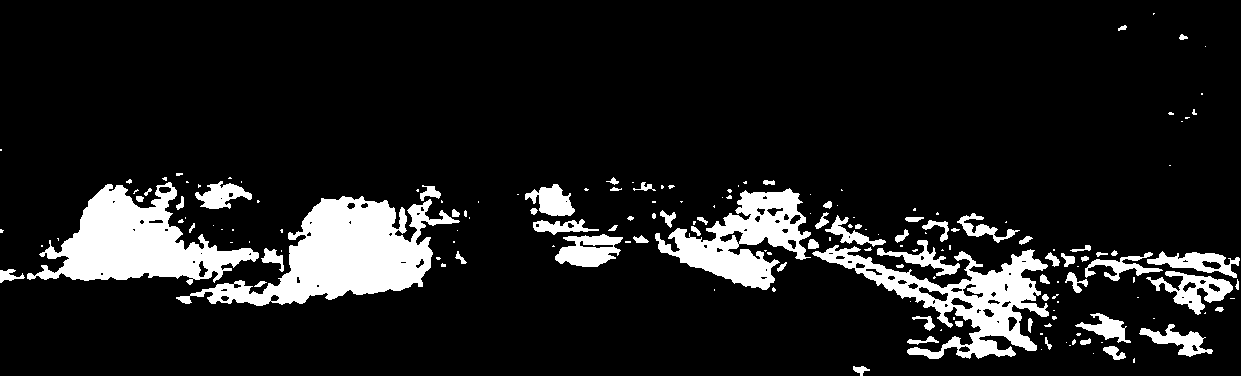}
        \caption{Segmentation at epoch 10.}
    \end{subfigure}
    ~ 
    \begin{subfigure}[b]{0.4\textwidth}
        \includegraphics[width=\textwidth]{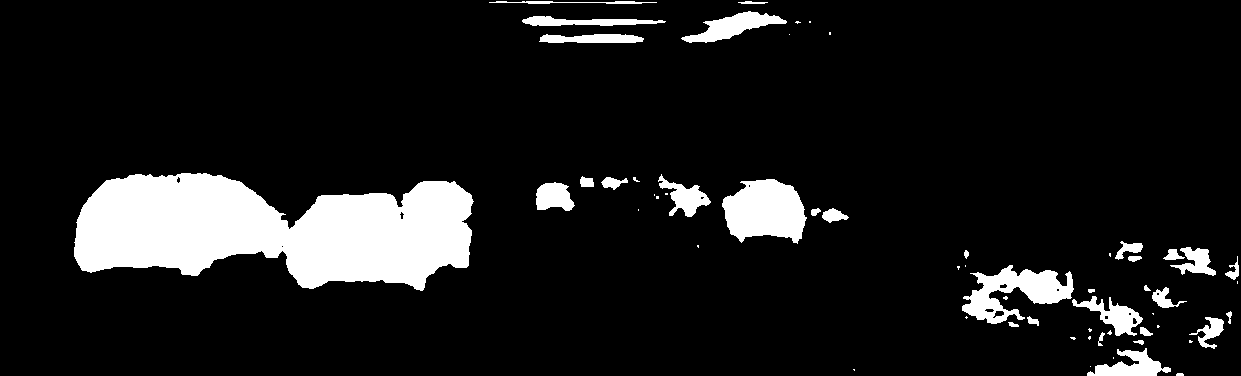}
        \caption{Segmentation at epoch 300.}
    \end{subfigure}
		\\
    \begin{subfigure}[b]{0.4\textwidth}
        \includegraphics[width=\textwidth]{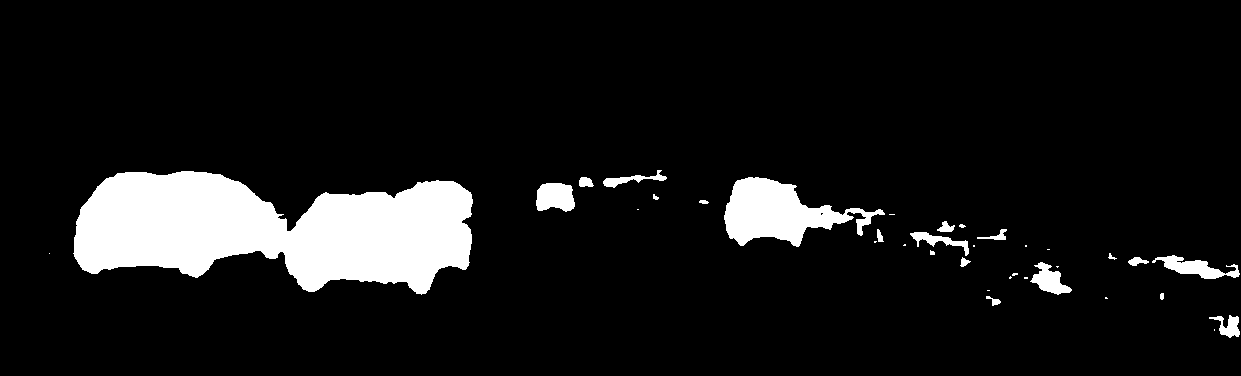}
        \caption{Segmentation at epoch 600.}
    \end{subfigure}
    \begin{subfigure}[b]{0.4\textwidth}
        \includegraphics[width=\textwidth]{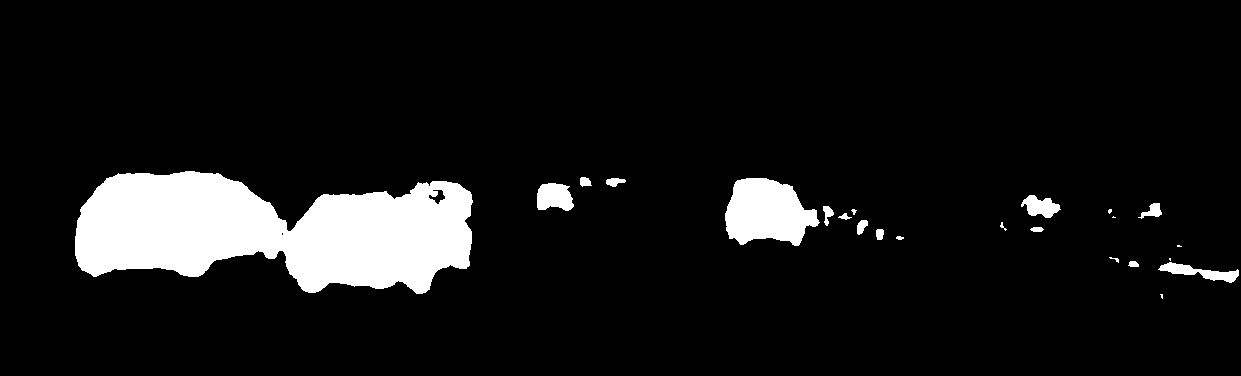}
        \caption{Segmentation at epoch 1000.}
    \end{subfigure}
    \caption{Coarse-to-fine foreground-background segmentation results using FBA-AMNet-32 models, at different training epochs, evaluated on a KITTI stereo 2015 test image.}\label{fig:segres}
\end{figure}

Figure~\ref{fig:segres} shows one image from the KITTI stereo 2015 test set and the coarse-to-fine foreground-background segmentation results generated by FBA-AMNet-32 models at training epoch $10$, $300$, $600$, and $1000$. The visualizations show that during the training process, the multitask network gradually learns better awareness of foreground objects. 
This shows how the network can learn the auxiliary task of foreground-background segmentation, while focusing more on learning the main task of disparity estimation.

\section{Conclusions \label{sec:conc}}

In this paper, we proposed atrous multiscale networks (AMNet) as a deep-learning based solution  to  the  problem of stereo disparity estimation. We proposed an atrous multiscale (AM) module that aggregates contextual features at multiple scales without the need for conventional downsampling and upsampling operations adopted by previous hour-glass modules. The AM module is used in feature extraction to aggregate the features extracted by our proposed depthwise separable residual network. We proposed an extended cost volume (ECV) to aggregate different disparity costs for a more accurate estimation. We also show how several AM modules can be stacked together with shortcut connections to form the stacked atrous multiscale (SAM) module which we use for fusion of the different volumes in the ECV and for cost aggregation at multiple scales. We also proposed the iterative multitask training of the foreground-background aware AMNet (FBA-AMNet) to learn the auxiliary task of foreground background segmentation for providing attention to the foreground-background transitions. Comparisons between FBA-AMNet and and AMNet throughout this paper confirm this benefit, and the FBA-AMNet also performs better than prior art that used class-based semantic segmentation. 
 Our method ranked first on the KITTI stereo 2015 leaderboard at the time we submitted our test results,  and performed better than previously published state-of-the-art methods on SceneFlow, KITTI stereo 2012, and Middlebury benchmarks most popular disparity estimation benchmarks.

In our future works, we plan  to deploy the proposed SAM networks for other tasks such as single-image depth estimation and semantic segmentation, as they showed a clear benefit over the previous state-of-art approaches.

{
\bibliographystyle{IEEEtranTCOM}
\bibliography{egbib}
}

\end{document}